\documentclass[10pt,twocolumn,letterpaper]{article}

\usepackage{3dv}
\usepackage{times}
\usepackage{epsfig}
\usepackage{graphicx}
\usepackage{amsmath}
\usepackage{amssymb}
\usepackage{enumerate}
\usepackage{paralist}
\usepackage{subfigure}

\threedvfinalcopy 

\def\threedvPaperID{51} 

\ifthreedvfinal\fi
\begin{document}

\title{MVS$^2$: Deep Unsupervised Multi-view Stereo with Multi-View Symmetry}

\author{Yuchao Dai, Zhidong Zhu, Zhibo Rao, Bo Li\\
School of Electronics and Information, Northwestern Polytechnical University, Xi'an, China \\ daiyuchao@gmail.com
}

\maketitle
\thispagestyle{empty}

\begin{abstract}
The success of existing deep-learning based multi-view stereo (MVS) approaches greatly depends on the availability of large-scale supervision in the form of dense depth maps. Such supervision, while not always possible, tends to hinder the generalization ability of the learned models in never-seen-before scenarios.
In this paper, we propose the first unsupervised learning based MVS network, which learns the multi-view depth maps from the input multi-view images and does not need ground-truth 3D training data. Our network is symmetric in predicting depth maps for all views simultaneously, where we enforce cross-view consistency of multi-view depth maps during both training and testing stages. Thus, the learned multi-view depth maps naturally comply with the underlying 3D scene geometry. Besides, our network also learns the multi-view occlusion maps, which further improves the robustness of our network in handling real-world occlusions. Experimental results on multiple benchmarking datasets demonstrate the effectiveness of our network and the excellent generalization ability.
\end{abstract}

\section{Introduction}
Multi-view stereo (MVS) targets at reconstructing the observed 3D scene structure from its multi-view images, whereas both the intrinsic calibration and extrinsic calibration between cameras are available. Traditional geometry-based approaches exploit multi-view photometric consistency and various kinds of regularizations/priors \cite{furukawa2009accurate}. Recently, the success of deep convolutional neural networks (CNNs) in monocular depth estimation \cite{li2018monocular,Unsupervised_Depth_Left_Right:CVPR_2017,Li_2015_CVPR} and binocular depth estimation \cite{zhong2017self,Zhong_2018_ECCV} has been extended to MVS. Existing deep CNNs based MVS approaches \cite{Yao2018MVSNet,Yao2019Recurrent,Huang2018DeepMVS,Sunghoon2019DSP} tend to represent MVS as an end-to-end regression problem. By exploiting large-scale ground truth 3D training data, these methods outperform traditional geometry-based approaches and dominate the leading boards on different benchmarking datasets \cite{Yao2018MVSNet, Yao2019Recurrent}. However, the success of these supervised MVS approaches strongly depends on the availability of large-scale ground-truth 3D training data, which not only not always available but also may further hinder their generalization ability in never-seen-before open-world scenarios \cite{Zhong_2018_ECCV}. Thus it is highly desired to develop unsupervised learning based MVS approaches. 

\begin{figure*}[htbp]
   \centering
   \includegraphics[width=0.85\textwidth]{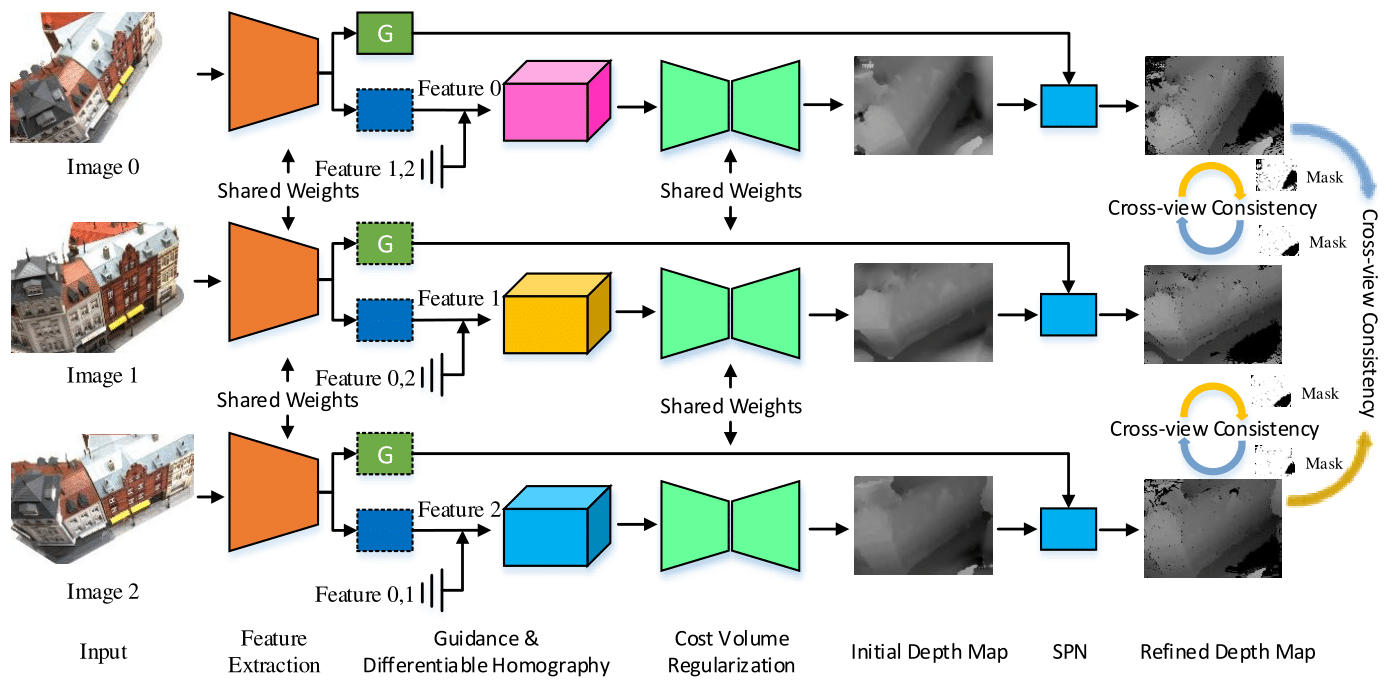} 
   \caption{\textbf{Our unsupervised deep multi-view stereo network architecture.} Our network consists of five modules, namely, feature extraction, guidance and differentiable homography, cost volume regularization, spn, and cross-view consistency loss evaluation. Note that, under our symmetric design, our network outputs consistent depth map for each image.}
   \label{fig:The overview of the Network architecture of this work}
\end{figure*}

In this paper, we propose the first unsupervised deep MVS network as shown in Fig.~\ref{fig:The overview of the Network architecture of this work}, which could be learned in an end-to-end manner and without using ground-truth depth maps as the supervision signals. We demonstrate that the multi-view image warping errors (photometric consistency across different views) themselves are sufficient to drive a deep network to converge to the correct state that leads to superior MVS performance. Our network structure differs from existing MVS and simple extension of unsupervised binocular stereo matching in the following aspects:
\begin{enumerate}[a)]
\setlength{\itemsep}{0pt}
\setlength{\parsep}{0pt}
\setlength{\parskip}{0pt}
\item Our network is \emph{symmetric} to all the views, \ie, it treats each view equivalently and predicts the depth map for each view simultaneously. Existing supervised learning based MVS methods \cite{Yao2018MVSNet,Yao2019Recurrent,Huang2018DeepMVS,wang2018mvdepthnet} apply an ``\emph{asymmetric}'' design and infer depth map for the reference image only. Thus, multiple depth maps estimated from different viewpoints do not comply with the same 3D geometry and 3D point clouds processing is required to derive a consistent 3D geometry. We would like to argue that this kind of ``centralized'' and ``asymmetric'' design has not fully exploited the multi-view relation encoded in the multi-view images.
\item We propose a new cross-view consistency in depth maps building upon our multi-view symmetry network design. The underlying principle is that as the multi-view images observe the 3D scene structure from different viewpoints, the estimated depth maps from MVS network should be consistent in 3D geometry. As our experiments demonstrate, this consistency plays a key role in strengthening the image warping error and guiding the network to coverage to meaningful states.
\item We integrate multi-view occlusion reasoning into our network, which enables us to detect occluded regions by using the cross-view consistency in depth maps. Under our framework, multi-view depth maps prediction and occlusion reasoning are alternatively updated.



\end{enumerate}

Our main contributions are summarized as follow:
\begin{enumerate}[1)]
\setlength{\itemsep}{0pt}
\setlength{\parsep}{0pt}
\setlength{\parskip}{0pt}
\item We present the first deep unsupervised MVS approach, which naturally fills the gap between traditional geometry-based approaches and deep supervised MVS methods. Our proposed unsupervised method avoids the necessity of large-scale 3D training data.
\item We introduce the cross-view consistency in depth maps and propose a loss function to measure the consistency. We demonstrate that this kind of consistency could be utilized to guide the training of a deep neural network.
\item Expensive experiments conducted on the SUN3D, RGB-D, DTU and Scenes11 benchmarking datasets demonstrate the effectiveness and the excellent generalization ability of our method. 
\end{enumerate}

\section{Related Work}
MVS has been an active research topic in geometric vision. Existing methods can be roughly classified into two categories: 1) Geometry-based MVS and 2) Supervised learning based MVS. We will also discuss related work in unsupervised monocular and binocular depth estimation.

\noindent\textbf{Geometry-based Multi-view Stereo:}
Traditional MVS methods focus on designing neighbor selection and photometric error measures for efficient and accurate reconstruction \cite{furukawa2015multi, Gallup2010Piecewise, Furukawa2010Towards}. Furukawa \etal \cite{Furukawa2009Manhattan} adopted geometric structures to reconstruct textured regions and applied Markov random fields to recover per-view depth maps. Langguth \etal \cite{Langguth2016Shading} used the shading-aware mechanism to improve the robustness of view selection. Wu \etal \cite{Wu2011High} utilized the lighting and shadows information to enhance the performance of the ill-posed region. Michael \etal \cite{Goesele2007Multi} chose images to match (both at a per-view and per-pixel level) for addressing the dramatic changes in lighting, scale, clutter, and other effects. Schonberger \etal \cite{Sch2016Pixelwise} proposed the COLMAP framework, which applied photometric and geometric priors to optimize the view selection and used geometric consistency to refine the depth map. 

\noindent\textbf{Supervised Deep Multi-view Stereo:}
Different from the above geometry-based methods, learning-based approaches adopt convolution operation which has powerful feature learning capability for better pair-wise patch matching \cite{Zhang2017Physically, Ji2017SurfaceNet, Kar2017Learning}. 
Ji \etal \cite{Ji2017SurfaceNet} pre-warped the multi-view images to 3D space, then used CNNs to regularize the cost volume. Huang \etal \cite{Huang2018DeepMVS} proposed DeepMVS, which aggregates information through a set of unordered images. Abhishek \etal \cite{Kar2017Learning} directly leveraged camera parameters as the projection operation to form the cost volume, and achieved an end-to-end network. Yao \etal \cite{Yao2018MVSNet} adopted a variance-based cost metric to aggregate the cost volume, then applied 3D convolutions to regularize and regress the depth map. Im \etal \cite{Sunghoon2019DSP} applied a plane sweeping approach to build a cost volume from deep features, then regularized the cost volume via a context-aware aggregation to improve depth regression. Very recently, Yao \etal \cite{Yao2019Recurrent} introduced a scalable MVS framework based on the recurrent neural network to reduce the memory-consuming. 

\noindent\textbf{Unsupervised Geometric Learning:}
Unsupervised learning has been developed in monocular depth estimation and binocular stereo matching by exploiting the photometric consistency and regularization. 
Xie \etal \cite{Xie2016Deep3D} proposed Deep3D to automatically convert 2D videos and images to stereoscopic 3D format. 
Zhou \etal \cite{Zhou2017Unsupervised} proposed an unsupervised monocular depth prediction method by minimizing the image reconstruction error. 
Mahjourian \etal \cite{Unsupervised_Depth_Motion_3D_Geometric_CVPR_2018} explicitly considered the inferred 3D geometry of the whole scene, where consistency of the estimated 3D point clouds and ego-motion across consecutive frames are enforced.
Zhong \etal \cite{zhong2017self,Zhong_2018_ECCV} used the image warping error as the loss function to derive the learning process for estimating the disparity map.






\section{Our Network}
In this section, we present our unsupervised learning based multi-view stereo network, MVS$^2$, which could be learned without the need of ground truth 3D data. We represent MVS as the task of predicting a depth map for each view simultaneously such that the estimated multiple depth maps comply with the underlying 3D geometry.
Our network structure follows the MVSNet model proposed in \cite{Yao2018MVSNet} but with significant modifications to achieve unsupervised MVS with multi-view symmetry, \ie, MVS$^2$. 

%
%
%


\subsection{Multi-view Symmetric Network Design}
Under the MVS configuration, each image observes the underlying 3D scene structure from different viewpoints. Therefore, the estimated depth maps from MVS network should be consistent in 3D geometry and each depth map estimation is not independent. However, existing deep MVS networks \cite{Yao2018MVSNet,Huang2018DeepMVS,Yao2019Recurrent} generally apply an ``asymmetric'' design and infer depth map for each image (termed as ``reference image'') individually. Thus, multiple depth maps estimated from different viewpoints do not necessarily comply with the same underlying 3D geometry. 

In this paper, we propose a de-centralized and multi-view symmetric network structure for MVS as illustrated in Fig.~\ref{fig:The overview of the Network architecture of this work}. Our network is \emph{symmetric} to all the views, \ie, it treats each view equivalently and predicts the depth map for each view simultaneously.
Our unsupervised deep MVS network consists of five modules, namely, multi-scale feature extraction, cost volume construction, cost volume regularization, depth map refinement through spatial propagation network, and unsupervised loss evaluation. 
We briefly describe each module with focus on how to achieve multi-view symmetry and how to enforce multi-view consistency.

\subsubsection{Cost Volume Reconstruction}
\begin{figure}[htbp]
   \centering
   \includegraphics[width=0.95\linewidth]{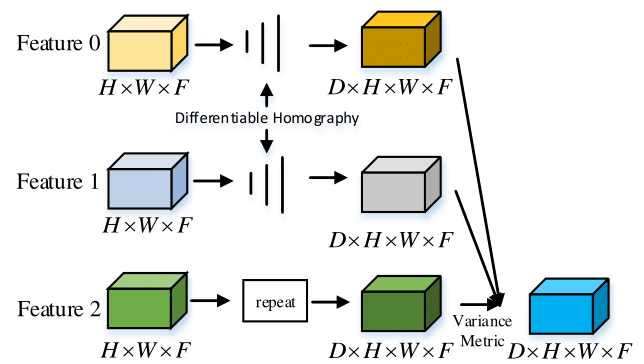} 
   \caption{\textbf{Cost Volume Construction.} The per-image cost volume is constructed by calculating the variance of the warped feature maps and the reference feature maps.}
   \label{fig:CostVolume}
\end{figure}

Under our multi-view symmetry configuration, we need to estimate a depth map for each input view. Following the MVSNet network, a cost volume has to be constructed for each input view. Denote the feature map extracted by feature extraction module for each view as $\mathcal{F}_i \in \mathbb{R}^{H\times W \times F}$, where $H,W,F$ denote the image height, image width and feature dimension correspondingly.
We adopt the classical plane sweeping based stereo pipeline and use differentiable homography matrix to warp the current image into each of the remaining images as shown in Fig.~\ref{fig:CostVolume}. 

In this way, we obtain $N-1$ warped feature volumes for each depth value $d$. We add the current feature volume into the group of warping feature volumes. Denote $D$ as the depth sample number, then we obtain $D$ groups of multiple feature volumes $\{V_{ij}\}_{j = 1, \cdots,N}$.
Finally, the multiple feature volumes are aggregated to one cost volume $\mathcal{C}_i \in \mathbb{R}^{D\times H \times W \times F}$ by using the variance operation \cite{Yao2018MVSNet}, which has been shown to be better than other operations such as mean or sum operation.




\subsubsection{Cost Volume Regularization}
\begin{figure}[htbp]
   \centering
   \includegraphics[width=0.95\linewidth]{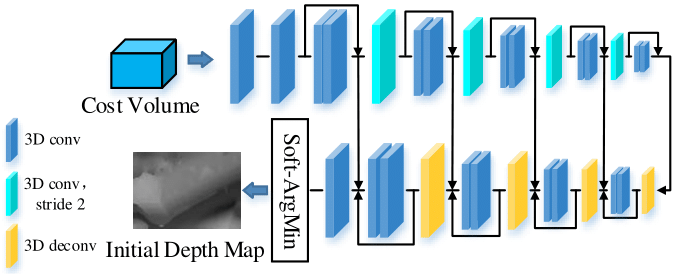} 
   \caption{\textbf{The cost volume regularization module}. It takes a cost volume as input, and is followed by a series of 3D CNNs.}
   \label{fig:3D-CNN}
\end{figure}

The raw cost volume $\mathcal{C}_i$ aggregated by the variance-based cost metric could be noise-contaminated, so we utilize 3D CNN to regularize each raw cost volume to generate a probability volume. After that, we apply the ArgMin operation to regress the depth map for the current view. The cost volume regularization process is illustrated in Fig.~\ref{fig:3D-CNN}.

As shown in Fig.~\ref{fig:3D-CNN}, we apply multi-scale 3D CNN to regularize the cost volume. The multi-scale 3D CNN consists of four-scale, where each convolutional operation is followed by a BN layer and a ReLU layer. On this base, we pass the feature maps between the same scale to form a residual architecture for avoiding losing the critical information. The output of our regularization module is a 1-channel volume $\mathcal{V}$ with dimension $D \times 1/4 H \times 1/4 W$.


Finally, we adopt the regression way to obtain an initial depth map $\mathcal{D}_{init}$. We first use the softmax function along the depth dimension to convert volume $\mathcal{V}$ to a probability map $\mathcal{P}$. Then, we apply the ArgMin operation to regress the depth map. The whole process is expressed as:
\begin{eqnarray}
\centering
\mathcal{D}_{init} = \sum\limits_{d=d_{min}}^{d_{max}}{d \times \mathcal{P} \left( d \right)} =  \sum\limits_{d=d_{min}}^{d_{max}}{d \times softmax \left( \mathcal{V}_d \right)},
\end{eqnarray}
where $d_{min}, d_{max}$ denote the min and max depth value. 

\subsubsection{Depth Map Refinement}
Even though the initial depth map is already a qualified output, the reconstruction boundaries of the object may suffer from over-smoothing due to up-sampling. To tackle this problem and improve the performance, we apply the spatial propagation network (SPN) \cite{liu2017learning} to refine the initial depth map. In this step, we obtain the guidance from the feature extraction module, and it could produce the affinity matrix which is spatially dependent on the input image. Then, we adopt the affinity matrix to guide the refinement process.

\subsection{Multi-view Occlusion reasoning}
\begin{figure}[htbp]
   \centering
   \includegraphics[width=0.95\linewidth]{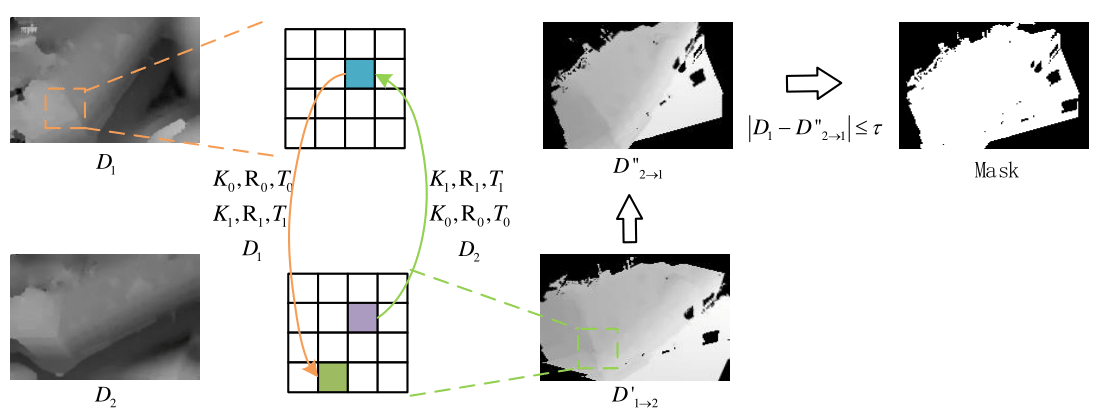} 
   \caption{\textbf{The cross-view depth consistency check process}. First, we warp the refined depth map $D_1$ to other view. Then, we re-warp the warping depth $D_{1\rightarrow2}^{'}$ map to current view. Final, we compare the refined depth map $D_1$ and warping depth map $D_{2\rightarrow1}^{''}$ to obtain the mask.}
   \label{fig:Mask}
\end{figure}

\begin{figure}[htbp]
\begin{center}
\begin{tabular}{c c c c}
\subfigure{
\includegraphics[width=0.19\linewidth]{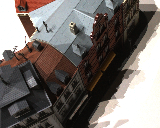}} &
\subfigure{
\includegraphics[width=0.19\linewidth]{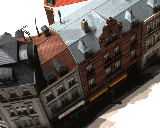}} &
\subfigure{
\includegraphics[width=0.19\linewidth]{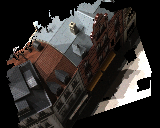}} &
\subfigure{
\includegraphics[width=0.19\linewidth]{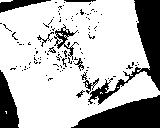}} \\
\scriptsize{(a) target img} & \scriptsize{(b) source img}  & \scriptsize{(c) warp img} & \scriptsize{(d) 5000 step} \\
\subfigure{
\includegraphics[width=0.19\linewidth]{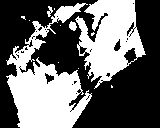}} &
\subfigure{
\includegraphics[width=0.19\linewidth]{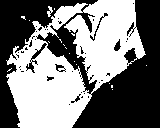}} &
\subfigure{
\includegraphics[width=0.19\linewidth]{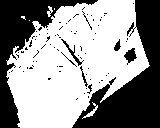}} &
\subfigure{
\includegraphics[width=0.20\linewidth]{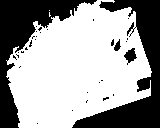}} \\
\scriptsize{(e) 50000 step} & \scriptsize{(f) 100000 step}  & \scriptsize{(g) 120000 step} & \scriptsize{(h) 150000 step} \\
\end{tabular}
\caption{\label{fig:The visualization and iteration of the occlusion mask.}\textbf{The visualization and iteration of the occlusion mask}. From left to right: target image, source image, warped image and the occlusion mask of various stages of the training process.(black means invalid points). Best Viewed on Screen.}
\end{center}
\end{figure}

Occlusion is inevitable to MVS, thus we have to decide the occlusion mask to avoid the occluded points from participating in the loss evaluation. 
Different from the occlusion mask detection based on forward-backward consistency check \cite{Sundaram2010Dense}\cite{Zou2018DF}, we exploit pixel-wise cross-view depth consistency to obtain the occlusion mask. Specifically, given a pair of estimated depth maps $D_i$ and $D_j$, we can synthesize two versions of $D_i$ by using the depth maps and the warping relations ${W}_{i\rightarrow j}$ and ${W}_{j\rightarrow i}$. The first order synthesized depth map ${D}_{i\rightarrow j}^{'}$ is generated by $D_i$ and ${W}_{i\rightarrow j}$. The second synthesized depth map ${D}_{j\rightarrow i}^{''}$ is generated by ${D}_{i\rightarrow j}^{'}$ and ${W}_{j\rightarrow i}$. 
The cross-view depth consistency check is illustrated in Fig.~\ref{fig:Mask}. 
Given perfect depth maps $D_i$ and $D_j$, ${D}_i$ and ${D}_{j\rightarrow i}^{''}$ should be the same up to occlusion. Therefore, we mark points which satisfy the constraint $|{D}_i$ - ${D}_{j\rightarrow i}^{''}| > \tau$ as invalid, where we set the threshold $\tau = 5$. 
For the sake of robustness, we use the cross-view depth consistency rather than the brightness consistency. 
In Fig.~\ref{fig:The visualization and iteration of the occlusion mask.}, we present a visualization of the evolution of the occlusion mask, its corresponding source images and synthesized image. It can be observed that with the increase of iterations the occlusion mask becomes more and more accurate.









\subsection{Loss Functions for Unsupervised MVS}
In this paper, we target to develop an unsupervised learning framework to estimate a fine and smooth depth map for each input image.
For optimizing the quality of depth map, we adopt two aspects of loss functions: view synthesis loss which includes unary term loss and smoothness loss, and cross-view consistency loss. Given multi-view images (${I}_1$,${I}_2$,...,${I}_n$), we first obtain the corresponding estimated depth maps during the training process. With the estimated depth map ${D}_i$ of the ${i}_{th}$ view and the given camera pose between ${i}_{th}$ and ${j}_{th}$ view (${I}_i$ and ${I}_j$), we can produce the synthesized view ${I}_{j\rightarrow i}^{'}$ of ${i}_{th}$ using the pixel of ${j}_{th}$ view and the mapping relations ${W}_{j\rightarrow i}$ between them. Similarly, we can also obtain the synthesized image ${I}_{i\rightarrow j}^{'}$ and the mapping relations ${W}_{i\rightarrow j}$. According to the bilateral mapping relations, we can produce the secondary synthesized image ${I}_{j\rightarrow i}^{''}$ using ${I}_{i\rightarrow j}^{'}$ and similarly the ${I}_{i\rightarrow j}^{''}$ using ${I}_{j\rightarrow i}^{'}$ with the bilinear sampler methods.

Our overall loss function can be formulated as follows:
\begin{eqnarray}
\centering
\mathcal{L} = \sum\limits_{i = 1}^{v -1} \sum_{
j=i+1}^{v}\mathcal{L}_{synthesis}^{i,j} +\mathcal{L}_{consistency},
\end{eqnarray}
where $v$ denotes the total amount of selected views. Apart from this, $\mathcal{L}_{synthesis}^{i,j}$ and $\mathcal{L}_{consistency}$ stand for the synthesized image loss between ${I}_i$ and ${I}_j$ and the cross-view consistency loss.

\subsubsection{View Synthesis Loss}
The view synthesis loss $\mathcal{L}_{synthesis}^{i,j}$ between ${I}_i$ and ${I}_j$ is defined as:
\begin{eqnarray}
\centering
\mathcal{L}_{synthesis}^{i,j} = \omega_u (\mathcal{L}_{u}^{i,j}+\mathcal{L}_{u}^{j,i}) + \omega_s \mathcal{L}_{s}^{i},
\end{eqnarray}
where $\mathcal{L}_{u}^{i,j}$ denotes the unary term loss and $\mathcal{L}_{s}^{i,j}$ denotes the depth field smoothness regularization loss.

\noindent\textbf{Unary term loss.} During the reconstruction process, we would like to minimize the discrepancy between the source image and the reconstructed image. Our loss consists of not only the $L_1$ distance between images and their gradients, but also the structure similarity SSIM. In order to further improve the robustness in brightness, we also exploit the Census transformation to measure the difference. Thus, our unary term loss is defined as follow:
\begin{equation}
\begin{split}
\mathcal{L}_{u}^{i,j} &= \frac1{|M|}\sum\left({  \lambda_1 \cdot \varphi\left({I}_i- I_{j\rightarrow i}^{'}\right)}  { +  \lambda_2 \cdot \varphi \left(\nabla I_i - \nabla I_{j\rightarrow i}^{'}\right) } \right. \\
& \left.{ +\lambda_3 \cdot \frac{1-\mathcal{S}(I_i, I_{j\rightarrow i}^{'})}{2}} { + \lambda_4 \cdot \varphi \left(C({I}_i)-C(I_{j\rightarrow i}^{'})\right)} \right)  \cdot{M},
\end{split}
\end{equation}
where $M$ is the unoccluded mask for obtaining the valid points. $\mathcal{S}(\cdot)$ denotes the structure similarity SSIM. $\varphi(s) = \sqrt{s^{2} + 0.001^{2}}$ can elevate the robustness of our loss. $\nabla(\cdot)$ denotes the gradient operator and $C{\cdot}$ denotes the Censum transformation of image. In this paper, we set $\lambda_1 = 0.5, \lambda_2 = 0.8, \lambda_3 = 0.5, \lambda_4 = 0.2$.

\noindent\textbf{Smoothness regularization term loss.}
To encourage the smoothness in the predicted depth map, the depth smoothness term is defined as:
\begin{equation}
\mathcal{L}_{s}^{i} = \frac{1}{N} \sum\left(e^{-\alpha_1\left|\nabla I_i\right|}\left|\nabla D_i\right|+e^{-\alpha_2\left|\nabla^2 I_i\right|}\left|\nabla^2 D_i\right|\right),
\end{equation}
where $\alpha_1 = 0.5, \alpha_2 = 0.5$. $N$ denotes the total number of the pixels.

\subsubsection{Cross-view Consistency Loss}
Besides the above brightness constancy loss, we also apply a new cross-view consistency loss by considering the consistency between the images and depth maps for these views. We introduce the following two losses: cross-view consistency loss $\mathcal{L}_{c}$ and multi-view brightness consistency loss $\mathcal{L}_{b}$,
\begin{eqnarray}
\centering
\mathcal{L}_{consistency}  = \sum\limits_{i = 1}^{v -1} \sum_{
j=i+1}^{v}(\mathcal{L}_{c}^{i,j} + \sum\limits_{k \neq j}^{v}\mathcal{L}_{b}^{i,j,k}).
\end{eqnarray}

The cross-view consistency loss consists of image consistency loss $\mathcal{L}_m$ based on images and depth consistency loss $\mathcal{L}_d$ based on depth maps. It can be formulated as:
\begin{equation}
\centering
\mathcal{L}_{c}^{i,j} = \lambda_5 \cdot (\mathcal{L}_{m}^{i,j} + \mathcal{L}_{m}^{j,i}) + \lambda_6 \cdot (\mathcal{L}_{d}^{i,j} + \mathcal{L}_{d}^{j,i}),
\end{equation}
where $\lambda_5 = 0.3, \lambda_6 = 0.3$.

Given two images ${I}_i$ and ${I}_j$, we can produce a synthesized image ${I}_{j\rightarrow i}^{'}$ by using ${I}_j$, ${D}_j$ and the relative pose between them. Naturally, we can also produce the secondary synthesized image ${I}_{i\rightarrow j}^{''}$ using ${I}_{j\rightarrow i}^{'}$, ${D}_i$ and their relative pose. Suppose that the predicted depth maps are accurate, then the discrepancy between ${I}_{i\rightarrow j}^{''}$ and ${I}_j$ should be very small and vice versa. In order to alleviate the robustness of the consistency loss, we also introduce another term to access the difference between ${I}_{i\rightarrow j}^{''}$ and ${I}_j$. The cross-view image consistency loss $\mathcal{L}_{m}$ is defined as:
\begin{equation}
\centering
\mathcal{L}_{m}^{i,j} = \mathcal{L}_{synthesis}({I}_j, {I}_{i\rightarrow j}^{''}).
\end{equation}
For the sake of robustness, we exploit the constraint between the predicted depth map ${D}_i$ and the synthesized depth map ${D}_{j\rightarrow i}^{'}$. Therefore, the cross-view depth consistency loss $\mathcal{L}_{d}$ is defined as:
\begin{equation}
\centering
\mathcal{L}_{d}^{i,j} = \varphi\left({D}_i- D_{j\rightarrow i}^{'}\right)\cdot{M},
\end{equation}


Besides the above consistency loss, we also present the multi-view brightness consistency loss to enhance the relationship of other views relative to the reference view. Our multi-view brightness consistency loss is formulated as:
\begin{equation}
\centering
\mathcal{L}_{b}^{i,j,k} = \mathcal{L}_{synthesis}({I}_{j\rightarrow i}^{''}, {I}_{k\rightarrow i}^{''}),
\end{equation}
which evaluates the brightness constancy across views $i,j,k$, \ie, multi-view consistency.

\begin{table*}[!t]
\scriptsize
\centering
\caption{\textbf{Quantitative results on the \textit{DTU}'s evaluation set \cite{Aan2016Large}}. We evaluate all methods using both the distance metric \cite{Aan2016Large} (lower is better), and the percentage metric \cite{Knapitsch2017Tanks} (higher is better) with $1mm$ thresholds}
\begin{tabular}{c c c c  c c c c c c c}
            \hline
           & \multicolumn{3}{c}{Accuracy(mm)}               & \multicolumn{3}{c}{Completeness(mm)} & overall  & \multicolumn{3}{c}{Percentage (\textless $1mm$)} \\ 
           & \multicolumn{3}{c}{Mean. Median.  Variance}     & \multicolumn{3}{c}{Mean. Median.  Variance} &   & \multicolumn{3}{c}{Acc.           Comp.           \textit{f-score}} \\ \hline
Camp \cite{Campbell2008Using}       & 0.835          & 0.482          & 1.549          &  0.554         & 0.523          & 4.076    & 0.695        & 71.75          & 64.94          & 66.31         \\ 
Furu \cite{furukawa2009accurate}       & 0.612          & 0.324          & 1.249          & 0.939          & 0.463          & 3.392     & 0.775           & 69.55          & 61.52          & 63.26         \\ 
Tola \cite{Tola2012Efficient}      & 0.343          & 0.210          & 0.385          & 1.190          & 0.492          & 5.319       & 0.766          & 90.49          & 57.83          & 68.07         \\ 
SurfaceNet\cite{Ji2017SurfaceNet} & 0.450          & 0.254           & 1.270          & 1.043       & 0.285          & 5.594      & 0.746          & 83.80          & 63.38          &           69.95          \\ 
MVSNet \cite{Yao2018MVSNet}     & 0.396     & 0.286   & 0.436   & 0.741    & 0.399    &  2.501  & 0.592    & 86.46      & 71.13     & 75.69 \\ \hline
MVS$^2$ (ours) & 0.760     & 0.485   & 1.791   & 0.515    & 0.307    &  1.121  & 0.637    & 70.56      & 66.12     & 68.27 \\
\hline
\end{tabular}%
\label{table:dtu}
\end{table*}

\section{Experimental Results}
To evaluate the performance of our proposed network MVS$^2$, we conducted experiments on widely used  multi-view stereo datasets, \eg DTU \cite{Aan2016Large}, SUN3D, RGBD, MVS and Scenes11 \footnote{https://github.com/lmb-freiburg/demon}. To align with other related works, we only trained our network on the training set of the DTU dataset, and directly tested on other datasets.



\subsection{Implementation Details}
\noindent\textbf{Dataset:} The DTU dataset\cite{Aan2016Large} is a large-scale multi-view stereo dataset, which consists of 128 scenes and each scene contains 49 images with 7 different lighting conditions. For a fair comparison, we follow the experimental setting in \cite{Yao2018MVSNet}. We generate the ground truth depth maps from the point cloud with the screened Poisson surface reconstruction method \cite{Kazhdan2013Screened}. We choose scenes: 1, 4, 9, 10, 11, 12, 13, 15, 23, 24, 29, 32, 33, 34, 48, 49, 62, 75, 77, 110, 114, 118 as the testing set and the other scenes as training set. 
The RGBD, SUN3D, MVS and Scenes11 datasets contain more than 30000 different scenes in total, which are very different from the DTU dataset. We use these datasets to validate the powerful generalization ability of our network.

\vspace{1mm}
\noindent\textbf{Training Details:}
Our MVS$^2$ network is implemented in Tensorflow with an NVIDIA v100 GPU. We train our model on the DTU's training set, but test it on the DTU's test set and other datasets directly. The image resolution for the DTU dataset is $640 \times 512$. The resolution of the predicted depth map is one-quarter of the original input due to down-sampling. The depth ranges are uniformly sampled from 425mm to 935mm with a resolution of 2.6mm and the depth sample number is $D = 192$.  For other datasets, in order to align the depth range, we set the depth start from 0.5mm with the depth sample resolution of 0.25, and the number of depth sample is $D=128$.

For the hyper-parameters, we set $\omega_w =0.8$, $\omega_s=0.1$ throughout the experiments. The batch size is set to 1 due to memory limit. The models are trained with RMSP optimizer for 10 epochs, with the learning rate of 2e-4 for the first 2 epochs and decreased by 0.9 for every two epochs.

\vspace{1mm}
\noindent\textbf{Error Metrics:} We use the standard metrics used in a public benchmark suite for performance evaluation. These quantitative measures include absolute relative error (Abs Rel), absolute difference error (Abs diff), square relative error (Sq Rel), root mean square error and its log scale (RMSE and RMSE log) and inlier ratio ($\delta < 1.25 ^ i$, $i=1,2,3$).

%

\subsection{Comparison with SOTA Methods}
\begin{figure}[htbp]
\begin{center}
\begin{tabular}{c c c}
\subfigure{
\includegraphics[width=0.28\linewidth]{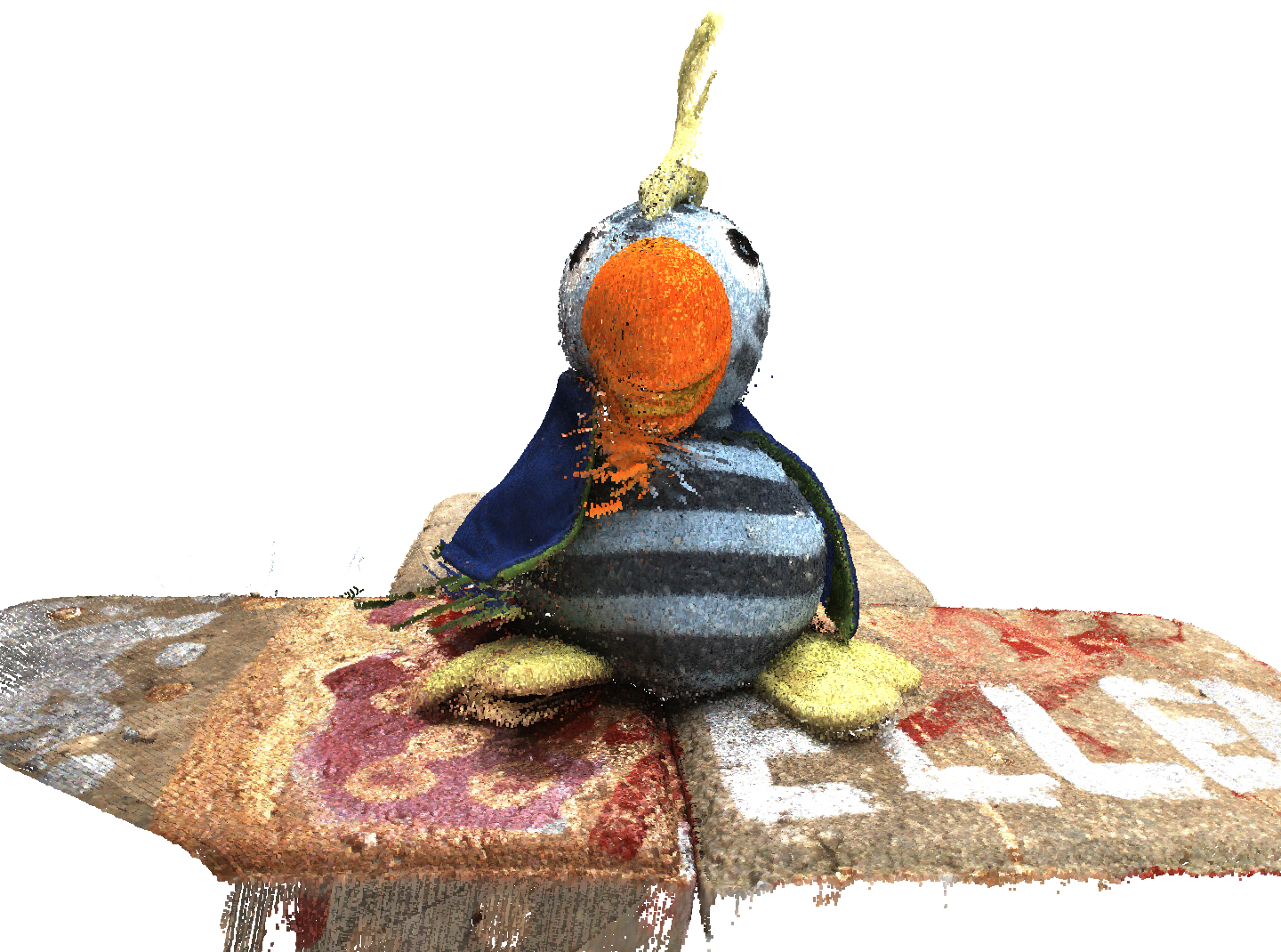}} &
\subfigure{
\includegraphics[width=0.28\linewidth]{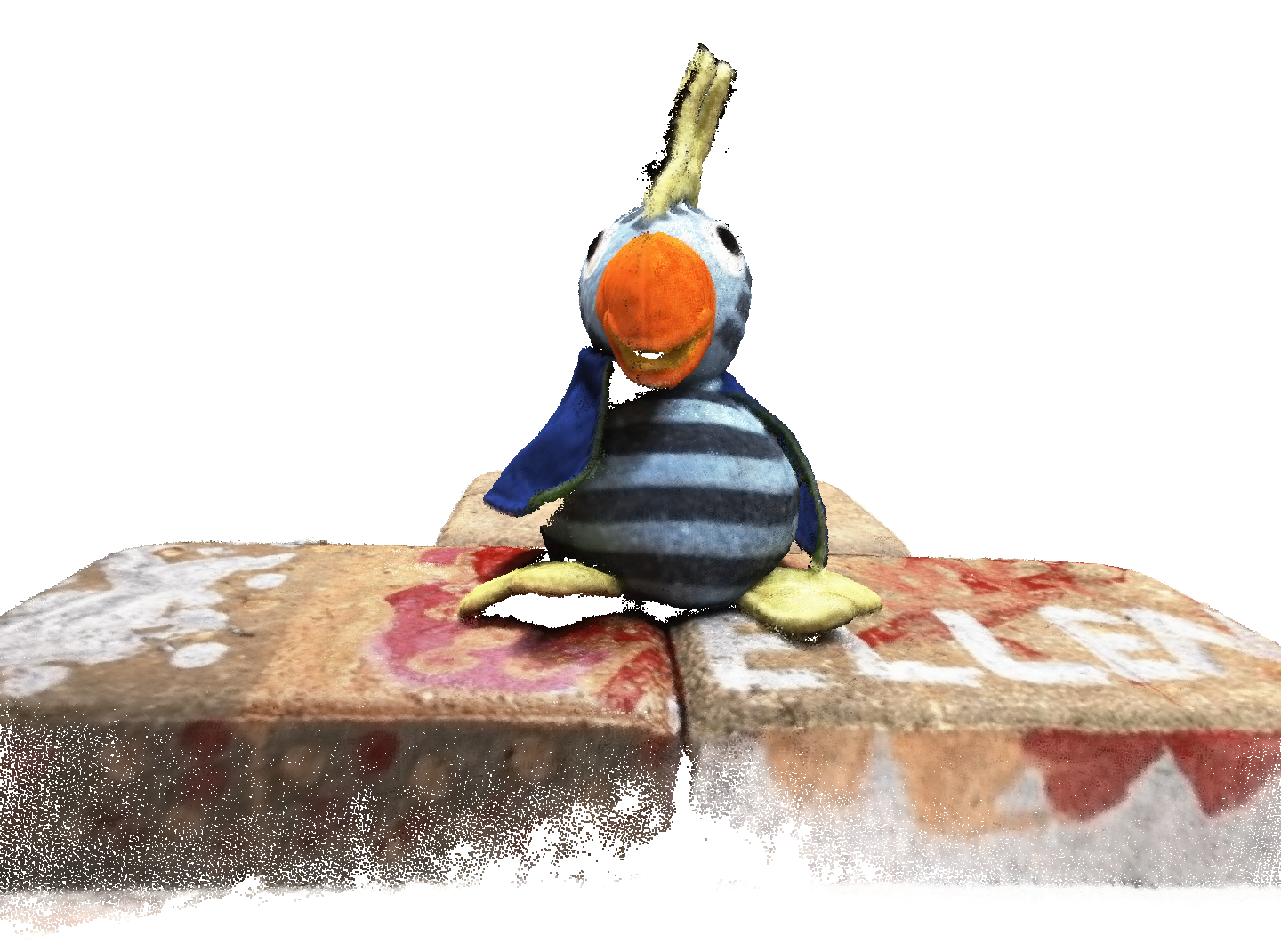}}&
\subfigure{
\includegraphics[width=0.28\linewidth]{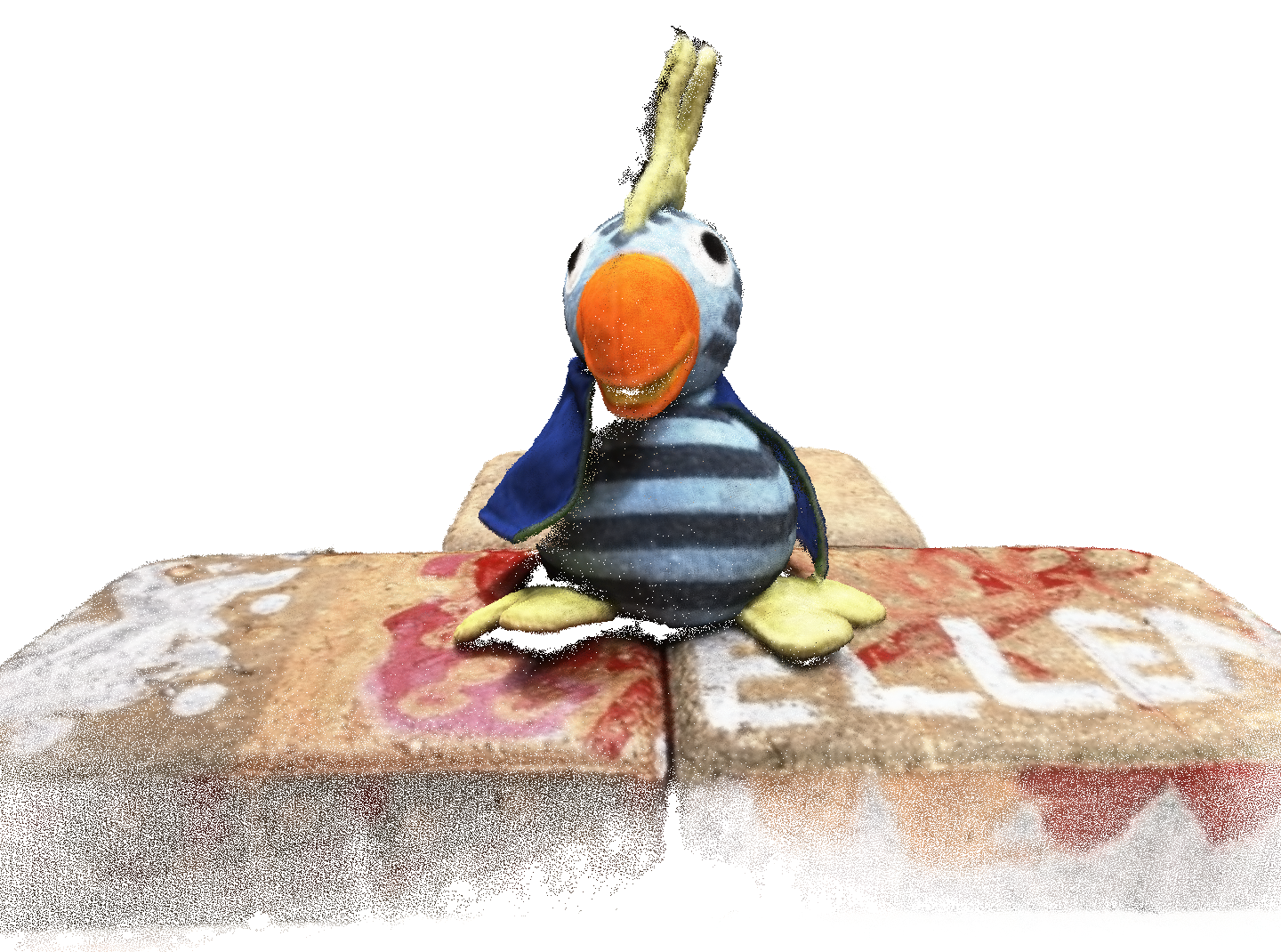}}\\
\subfigure{
\includegraphics[width=0.28\linewidth]{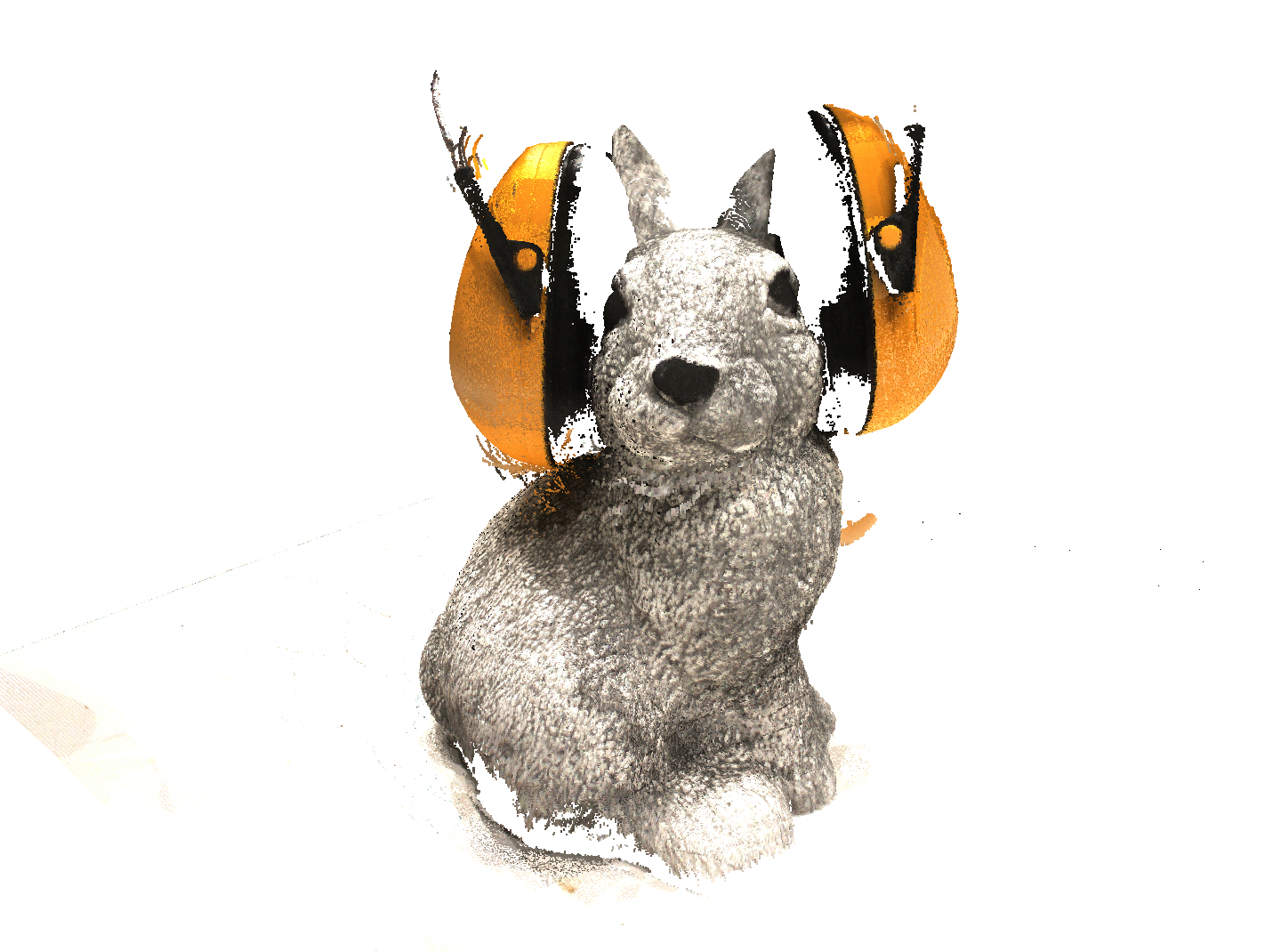}}&
\subfigure{
\includegraphics[width=0.28\linewidth]{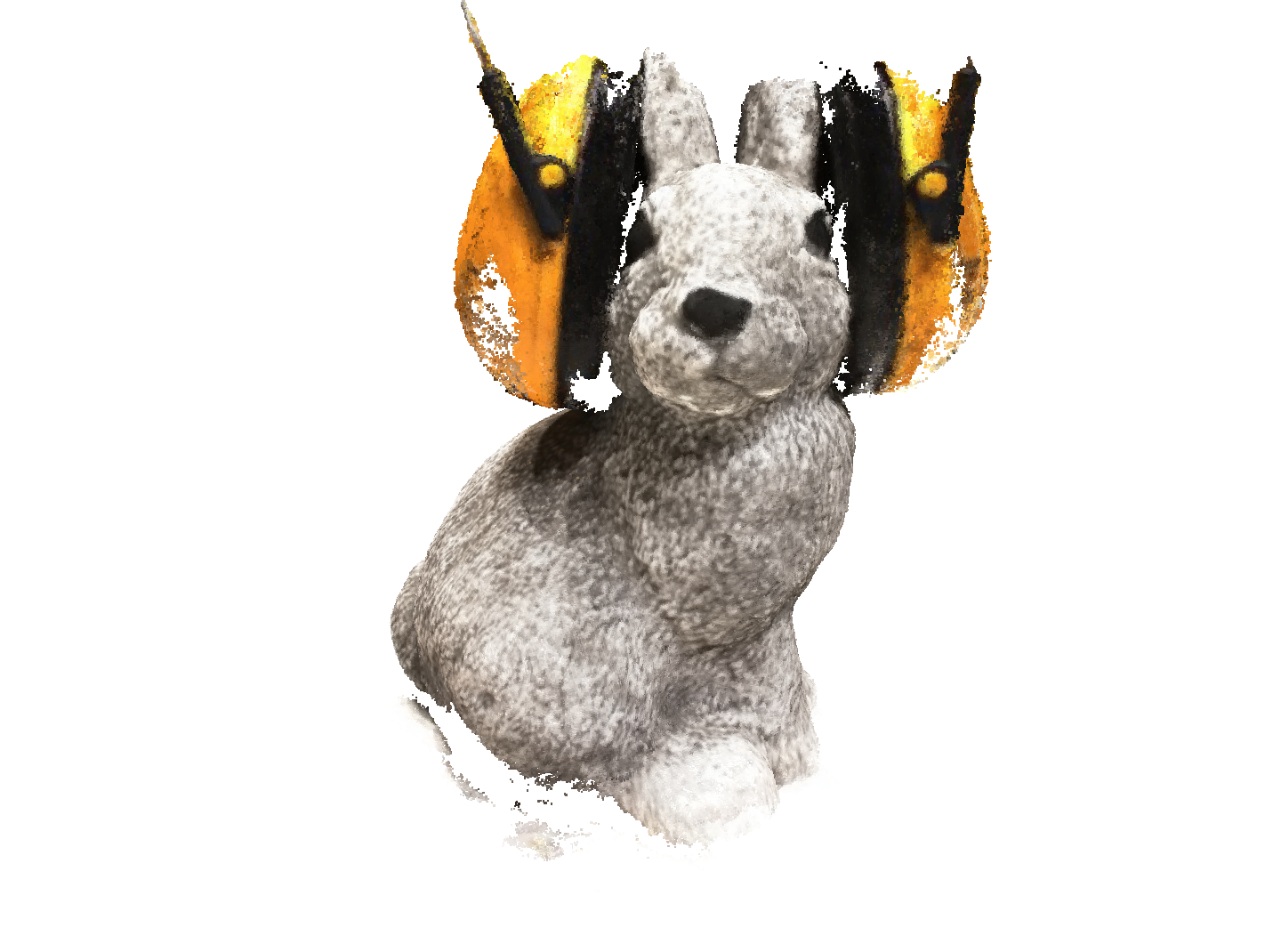}} &
\subfigure{
\includegraphics[width=0.28\linewidth]{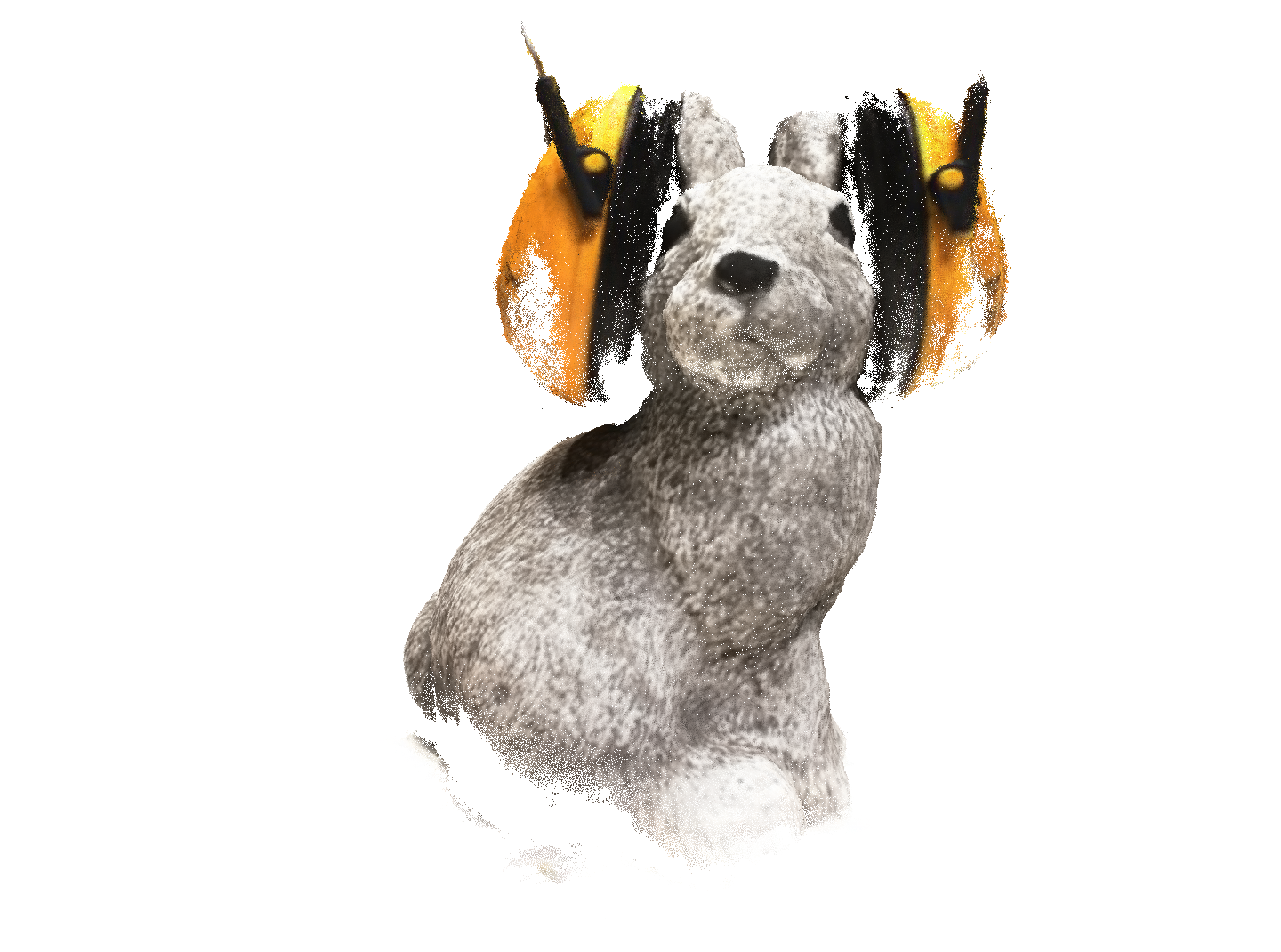}} \\
\subfigure{
\includegraphics[width=0.28\linewidth]{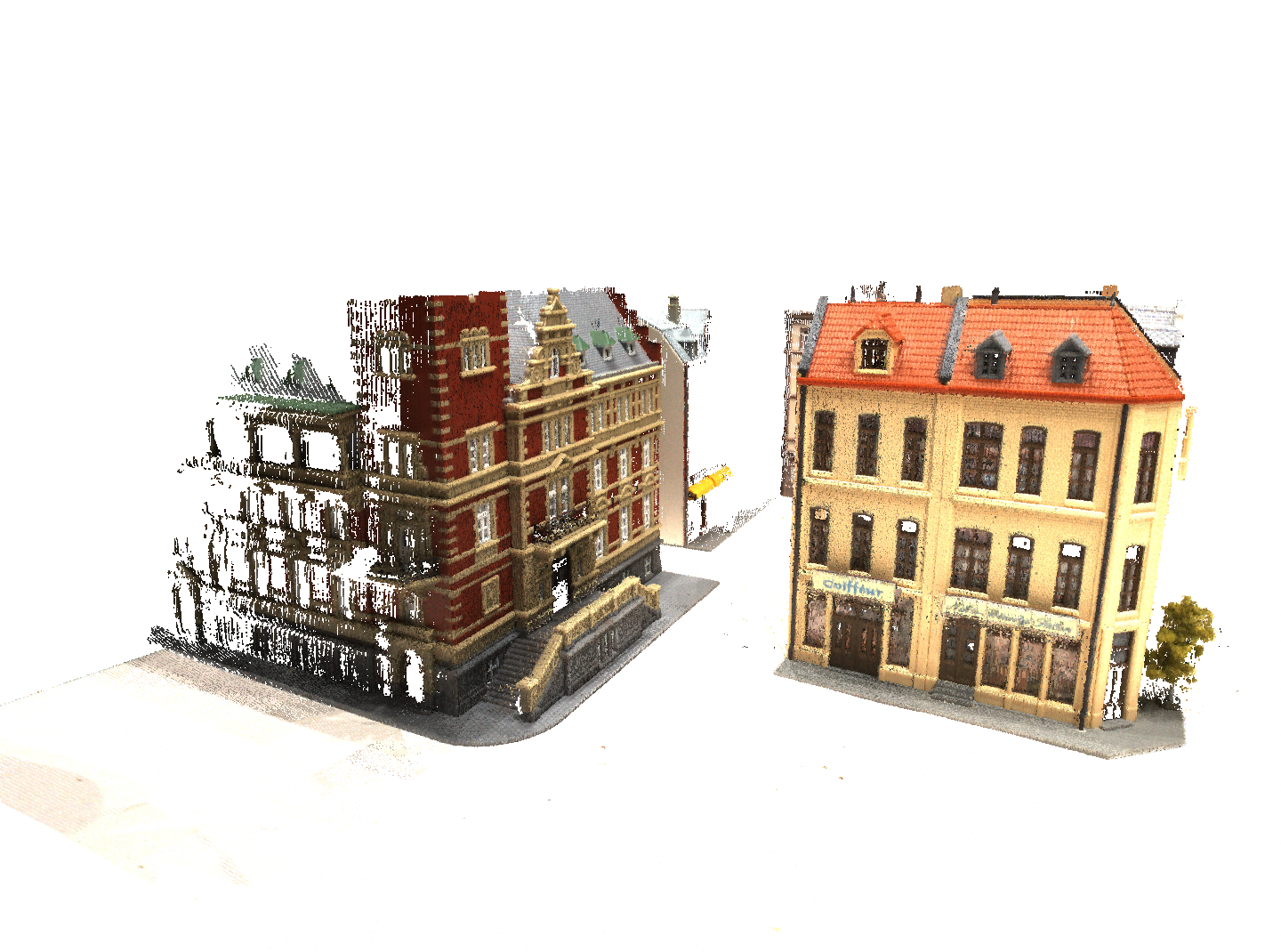}} &
\subfigure{
\includegraphics[width=0.28\linewidth]{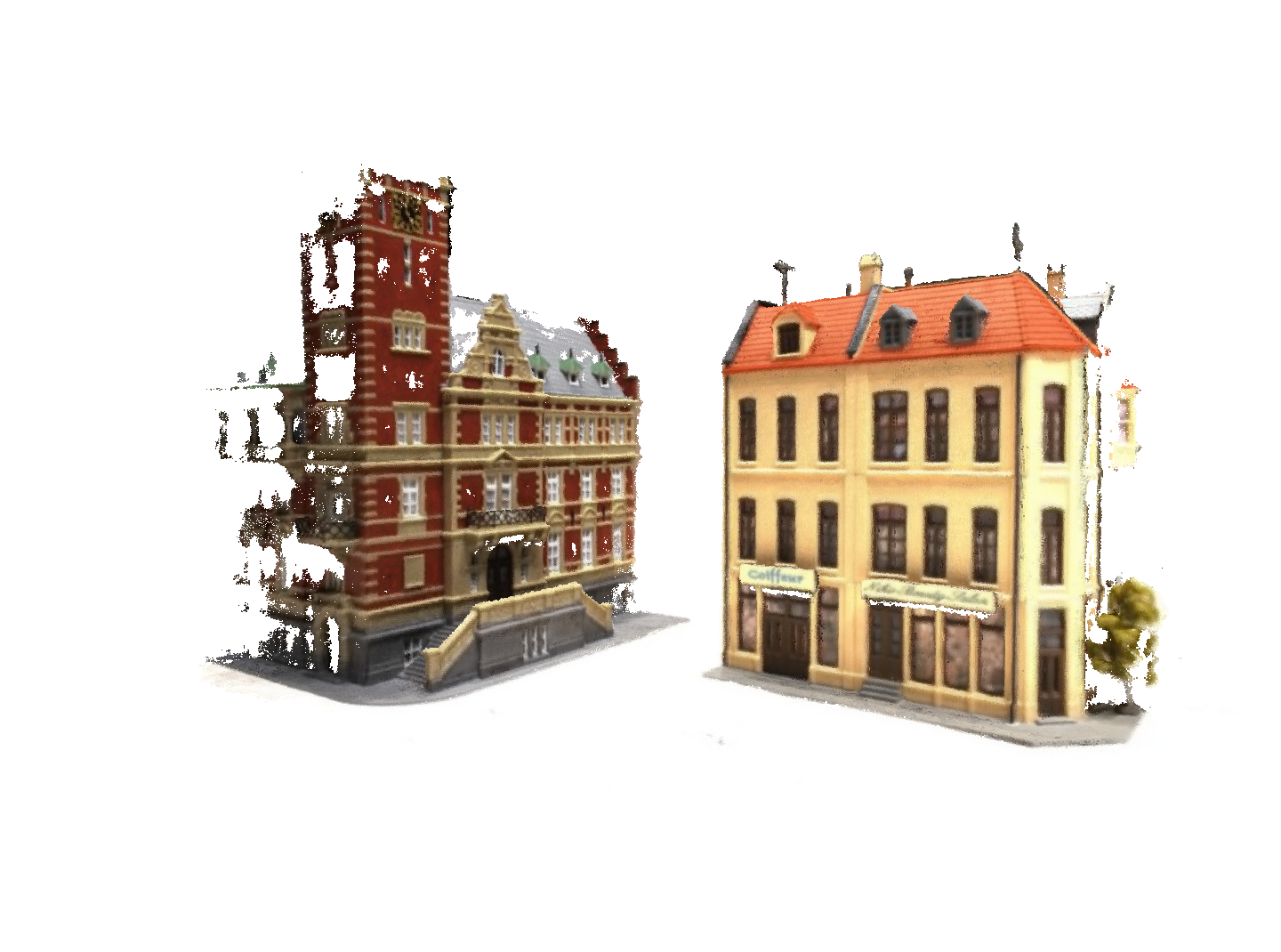}} &
\subfigure{ 
\includegraphics[width=0.28\linewidth]{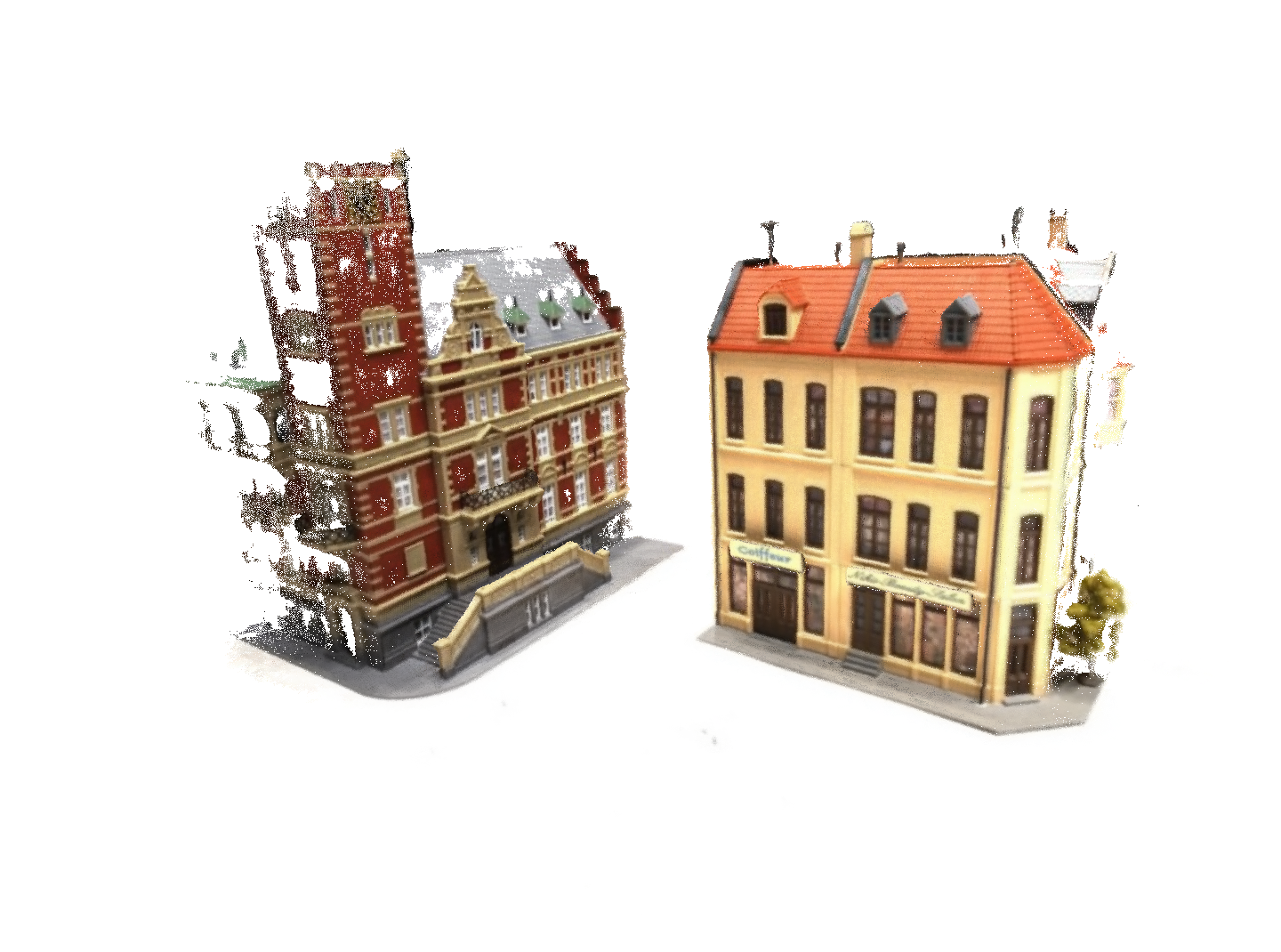}} \\
\small{(a) } Ground Truth & \small{(b) Yao's result \cite{Yao2018MVSNet}}  & \small{(c) Our result} \\
\end{tabular}
\caption{\label{fig:The qualitative results of our MVS$^2$ on DTU dataset.}\textbf{Qualitative comparison in 3D reconsturciton between our MVS$^2$ and SOTA supervised MVS method \cite{Yao2018MVSNet} on the DTU dataset.} From left to right: ground-truth point clouds, Yao's point clouds \cite{Yao2018MVSNet}, our point clouds. Best Viewed on Screen.}
\end{center}
\end{figure}

To verify the performance of our MVS$^2$, we tested it on the widely used DTU dataset. First, we conducted extensive quantitative comparisons with the state-of-the-art (SOTA) methods published recently. Performance comparison with other SOTA MVS methods is reported in Tab.~\ref{table:dtu}. From Tab.~\ref{table:dtu}, we can conclude that MVS$^2$ achieves higher completeness than other SOTA MVS methods while achieving comparable performance under other metrics. 
We applied a depth map fusion step to integrate the depth maps from different views to a unified point cloud representation. We chose the gipuma \cite{Galliani2015Massively} to fuse our depth maps.
The qualitative comparisons in 3D reconstruction are shown in Fig.~\ref{fig:The qualitative results of our MVS$^2$ on DTU dataset.}, where MVS$^2$ achieves 3D reconstruction comparable with state-of-the-art supervised MVS method \cite{Yao2018MVSNet}. 

\begin{table*}[htbp]
\footnotesize
\centering
\caption{\textbf{Ablation Experiments}. (a) Without spatial propagation refine module, only compute self-supervised loss through initial depth map generated by inference network. (b) With cost volume generated just by the variance-based homography feature. (c) Without the view consistency self-supervised loss. (d) Our complete MVS$^2$. Datasets: DTU dataset}
\begin{tabular}{l ccccc ccc c} 
\hline
& & \multicolumn{4}{c}{Error metric} & \multicolumn{3}{c}{Accuracy metric({$\delta < \alpha ^ i$})} \\ 
w/o  & Abs Rel   & Abs Diff   & Sq Rel  & RMSE   & RMSE log   & $\alpha$  & $\alpha ^ 2$ & $\alpha ^ 3$  &runtime  \\ \hline
(a)SPN Refine 	& 0.0175 & 13.0339 & 1.8440 & 30.4543 & 0.0187 & 0.9814 & 0.9992 & 1.0000 & 0.313s \\ \hline
(b)Cost(difference)   	& 0.0204   & 15.1751 & 2.3806 & 34.6147 & 0.0251 & 0.9753 & 0.9986 & 1.0000 & 0.273s \\
 \hline
(c)view consistency  	& 0.0355   & 24.9464 & 5.2399 & 55.4236 & 0.0425  & 0.9482 & 0.9920 & 0.9998 & 0.322s \\
 \hline
 (d)MVS$^2$ (ours) 	& 0.0147 & 11.3912 & 1.5478 & 28.4428 & 0.0156 & 0.9900 & 1.0000 & 1.0000 & 0.325s \\
 \hline
      \end{tabular}
\label{tab:Ablation Experiments}
\end{table*}

\begin{table}[htbp]
\scriptsize
\centering
\caption{\textbf{Ablation experiments with different combinations of consistency methods}. (a) With only the multi-view brightness consistency (BC) loss. (b) With only the cross-view consistency check (CC) loss. (c) With all loss. Datasets: DTU dataset}
\begin{tabular}{l ccccc} \hline
& & \multicolumn{4}{c}{Error metric} \\ 
& Abs Rel   & Abs Diff   & Sq Rel  & RMSE   & RMSE log   \\ \hline
(a) BC  	& 0.0180 & 13.4363 & 1.7042 & 30.2826 & 0.0190  \\ \hline
(b) CC   	& 0.0172 & 12.8649 & 1.6311 & 29.4134 & 0.0182 \\
 \hline
(c) BC and CC  	& 0.0147 & 11.3912 & 1.5478 & 28.4428 & 0.0156 \\
 \hline
      \end{tabular}
\label{tab:Ablation experiments with different combinations of consistency methods}
\end{table}

\subsection{Ablation Studies}
To analyze the contribution of different modules of our network model, we conduct three ablation studies on the DTU validation set with $W \times H \times D = 640 \times 512 \times 192$. Quantitative results are reported in Tab. ~\ref{tab:Ablation Experiments}.

\noindent\textbf{SPN Refinement.}
Under our network model, we introduce the spatial propagation network (SPN) \cite{liu2017learning} to refine the initial depth map. To analyze the contribution of this module, we conduct experimental comparison with and without this module and the results are reported in Tab.~\ref{tab:Ablation Experiments}.
It can be observed that when the SPN module is removed, the performance consistently drops. For example the Abs Diff increases from 11.3912 to 13.0339, and the Abs Rel increases from 0.0147 to 0.0175, which clearly demonstrates the effectiveness of the SPN refinement module.

\noindent\textbf{Cost Volume.}
In building the cost volume, we exploit both the feature for the current view and the variance-based feature \cite{Yao2018MVSNet}. To validate the effectiveness of our cost volume construction, we compare with a baseline implementation by using the variance-based feature only, which is used in \cite{Yao2018MVSNet}. As illustrated in Tab.~\ref{tab:Ablation Experiments}, when the feature for the current view is excluded from the cost volume, the performance consistently drops. For example the Abs Rel jumps from 0.0147 to 0.0204 and the Abs Diff increases from 11.3912 to 15.1751. The experimental results prove the effectiveness of our proposed cost volume reconstruction method in exploiting the feature of the current view.

\noindent\textbf{Consistency Loss.}
In this paper, we have proposed a consistency loss to further constrain the multiple estimated depth maps, which is also a key contribution. To analyze the contribution of this consistency loss, we conducted experiments with and without this loss term and the results are reported in Tab.~\ref{tab:Ablation Experiments}. When the cross-view consistency loss is removed from our unsupervised loss, the performance deteriorates sharply. For example the Abs Rel shoots up from 0.0147 to 0.0355 while the Abs Diff increases from 11.3912 to 24,9464 and the Sq Rel jumps from 1.5478 to 5.2399. The experimental results clearly demonstrate the significance of our proposed consistency loss.

Besides the above ablation studies in analyzing the contribution of our novel consistency loss term, as our consistency term actually consists of two terms (multi-view brightness consistency and cross-view consistency in depth maps), we also conducted two additional experiments to analyze the effectiveness of each term and the corresponding results are reported in Tab.~\ref{tab:Ablation experiments with different combinations of consistency methods}. From Tab.~\ref{tab:Ablation experiments with different combinations of consistency methods}, we could draw the following conclusions that: 1) Both the multi-view brightness consistency term (BC) and the cross-view consistency term (CC) are critical for achieving improved performance; 2) The cross-view consistency term (CC) plays a more important role than the multi-view brightness consistency term (BC) in depth map estimation.

\begin{table*}[!t]
{
\footnotesize
\centering
\caption{\textbf{Generalization Ability}. Multi-view stereo methods: COLMAP, DeepMVS, DeMoN, where Deep MVS and DeMoN are supervised methods and trained on these datasets correspondingly.}
\begin{tabular}{cc ccccc ccc} 
\hline
& & \multicolumn{5}{c}{Error metric} & \multicolumn{3}{c}{Accuracy metric({$\delta < \alpha ^ i$})} \\ 
Datasets & Method  & Abs Rel   & Abs Diff   & Sq Rel  & RMSE   & RMSE log   & $\alpha$  & $\alpha ^ 2$ & $\alpha ^ 3$    \\ 
\hline
SUN3D & COLMAP\cite{Sch2016Pixelwise} 	& 0.6232 & 1.3267 & 3.2359 & 2.3162 & 0.6612 & 0.3266 & 0.5541 & 0.7180 \\
      & DeMoN\cite{Ummenhofer2017DeMoN} 	& 0.2137 & 2.1477 & 1.1202 & 2.4212 & 0.2060 & 0.7332 & 0.9219 & 0.9626 \\
      & DeepMVS\cite{Huang2018DeepMVS} & 0.2816 & 0.6040 & 0.4350 & 0.9436 & 0.3633 & 0.5622 & 0.7388 & 0.8951 \\
      & MVS$^2$ (ours)   & 0.3488 & 0.5956 & 0.4879 & 0.7525 & 0.3805 & 0.4930 & 0.7616 & 0.9100 \\ \hline
RGBD  & COLMAP \cite{Sch2016Pixelwise} 	& 0.5389 & 0.9398 & 1.7608 & 1.5051 & 0.7151 & 0.2749 & 0.5001 & 0.7241 \\
      & DeMoN \cite{Ummenhofer2017DeMoN}	& 0.1569 & 1.3525 & 0.5238 & 1.7798 & 0.2018 & 0.8011 & 0.9056 & 0.9621 \\
      & DeepMVS \cite{Huang2018DeepMVS} & 0.2938 & 0.6207 & 0.4297 & 0.8684 & 0.3506 & 0.5493 & 0.8052 & 0.9217 \\
      & MVS$^2$ (ours)   & 0.4414 & 0.8698 & 0.9352 & 1.2853 & 0.4726 & 0.4657 & 0.6878 & 0.8057 \\ \hline
MVS   & COLMAP \cite{Sch2016Pixelwise}	& 0.3841 & 0.8430 & 1.257 & 1.4795 & 0.5001 & 0.4819 & 0.6633 & 0.8401 \\
      & DeMoN\cite{Ummenhofer2017DeMoN} 	& 0.3105 & 1.3291 & 19.970 & 2.6065 & 0.2469 & 0.6411 & 0.9017 & 0.9667 \\
      & DeepMVS \cite{Huang2018DeepMVS} & 0.2305 & 0.6628 & 0.6151 & 1.1488 & 0.3019 & 0.6737 & 0.8867 & 0.9414 \\
      & MVS$^2$ (ours)   & 0.3729 & 0.8170 & 0.9135 & 1.3938 & 0.4921 & 0.5136 & 0.6952 & 0.9123 \\\hline
SCENES11 & COLMAP\cite{Sch2016Pixelwise} 	& 0.6249 & 2.2409 & 3.7148 & 3.6575 & 0.8680 & 0.3897 & 0.5674 & 0.6716 \\
         & DeMoN \cite{Ummenhofer2017DeMoN}	& 0.5560 & 1.9877 & 3.4020 & 2.6034 & 0.3909 & 0.4963 & 0.7258 & 0.8263 \\
         & DeepMVS \cite{Huang2018DeepMVS} & 0.2100 & 0.5967 & 0.3727 & 0.5909 & 0.2699 & 0.6881 & 0.8940 & 0.9687 \\
         & MVS$^2$ (ours)    & 0.5981 & 2.0848 & 3.3365 & 2.9477 & 0.4885 & 0.4695 & 0.6531 & 0.7879 \\ \hline
\end{tabular}

\label{tab:The Excellent Generalization Ability}
}
\end{table*}

\begin{figure*}
\begin{center}
\begin{tabular}{c c c c}
\subfigure{
\includegraphics[width=0.2\linewidth]{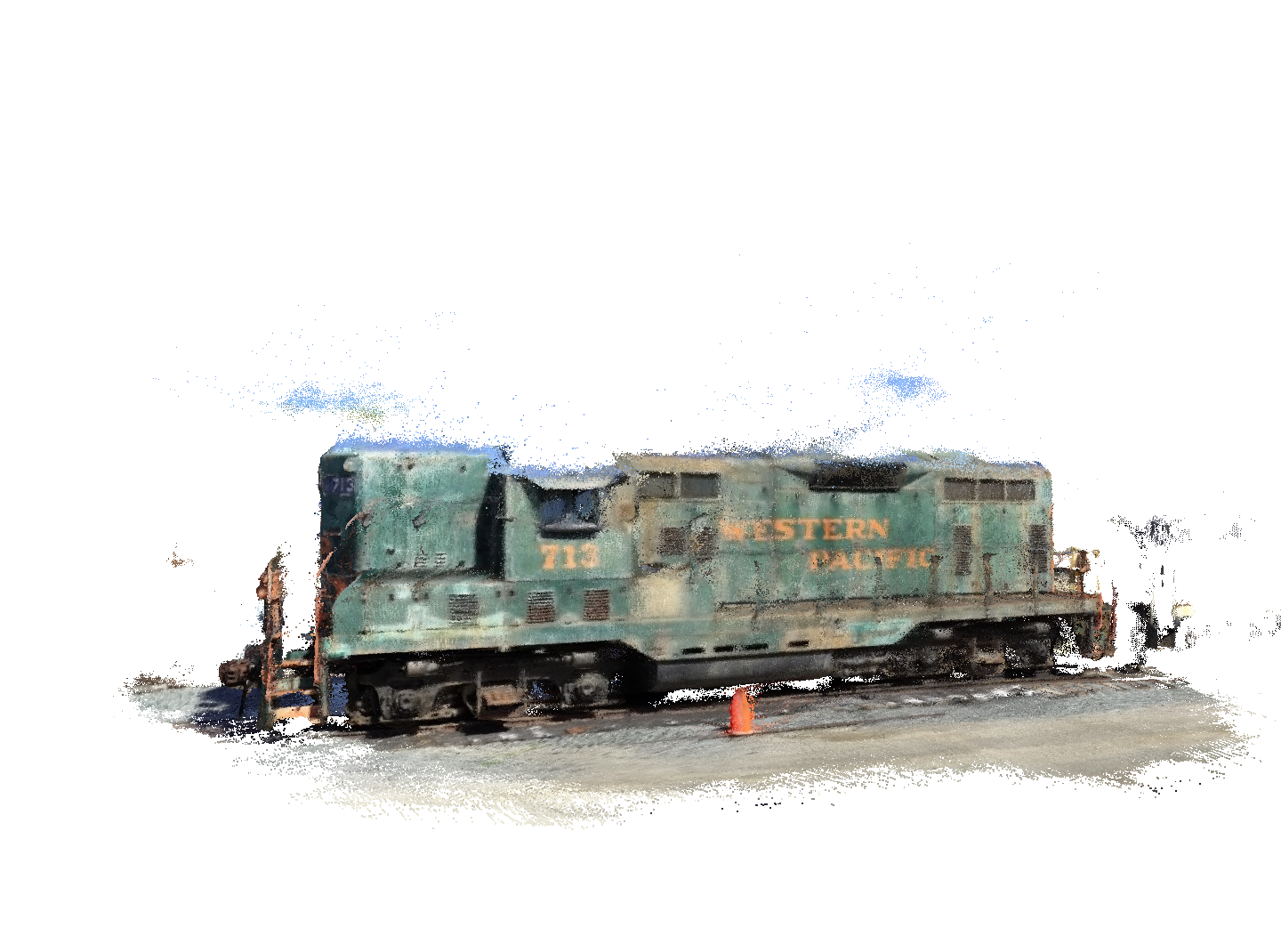}} &
\subfigure{
\includegraphics[width=0.2\linewidth]{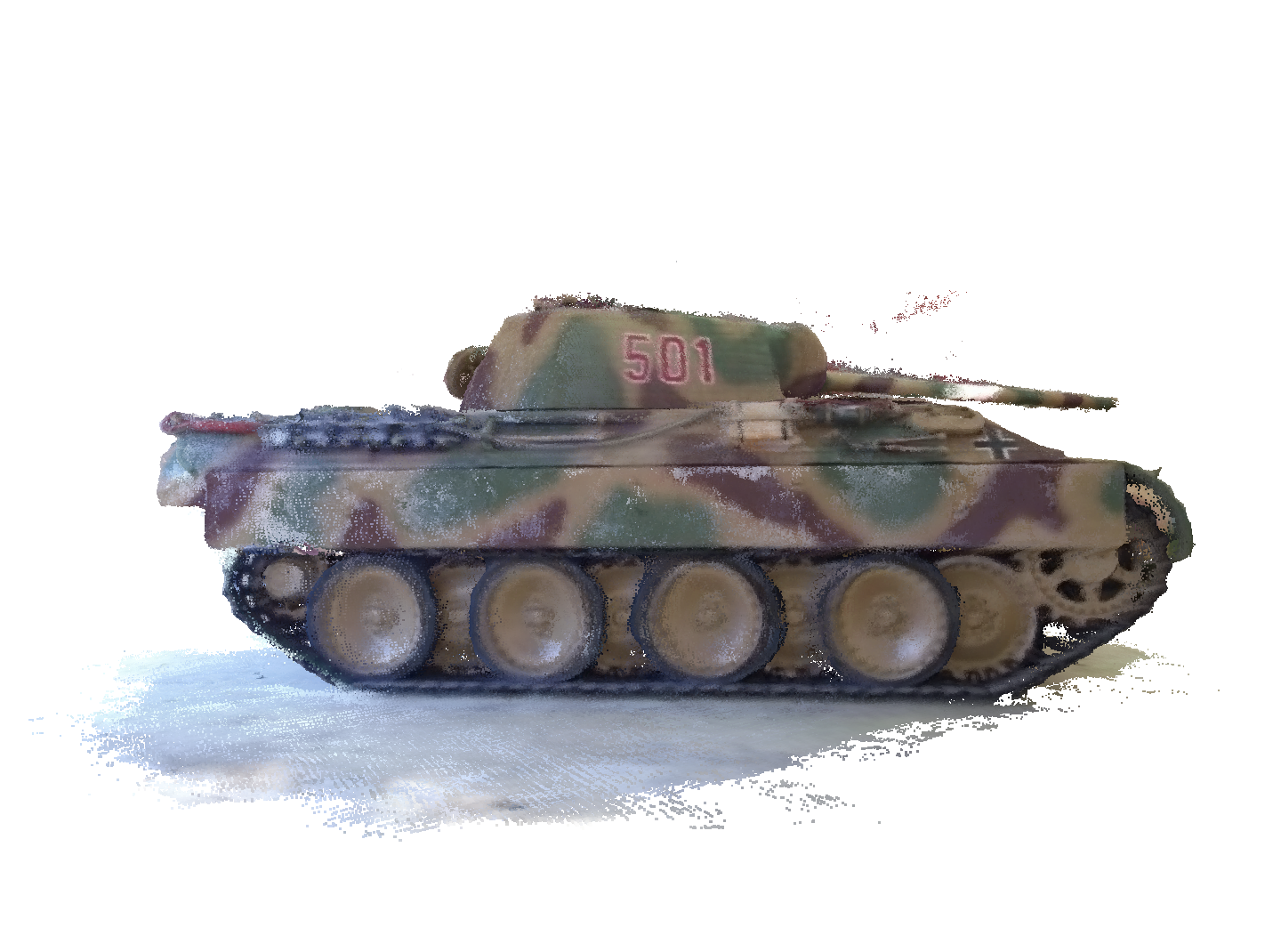}} &
\subfigure{
\includegraphics[width=0.2\linewidth]{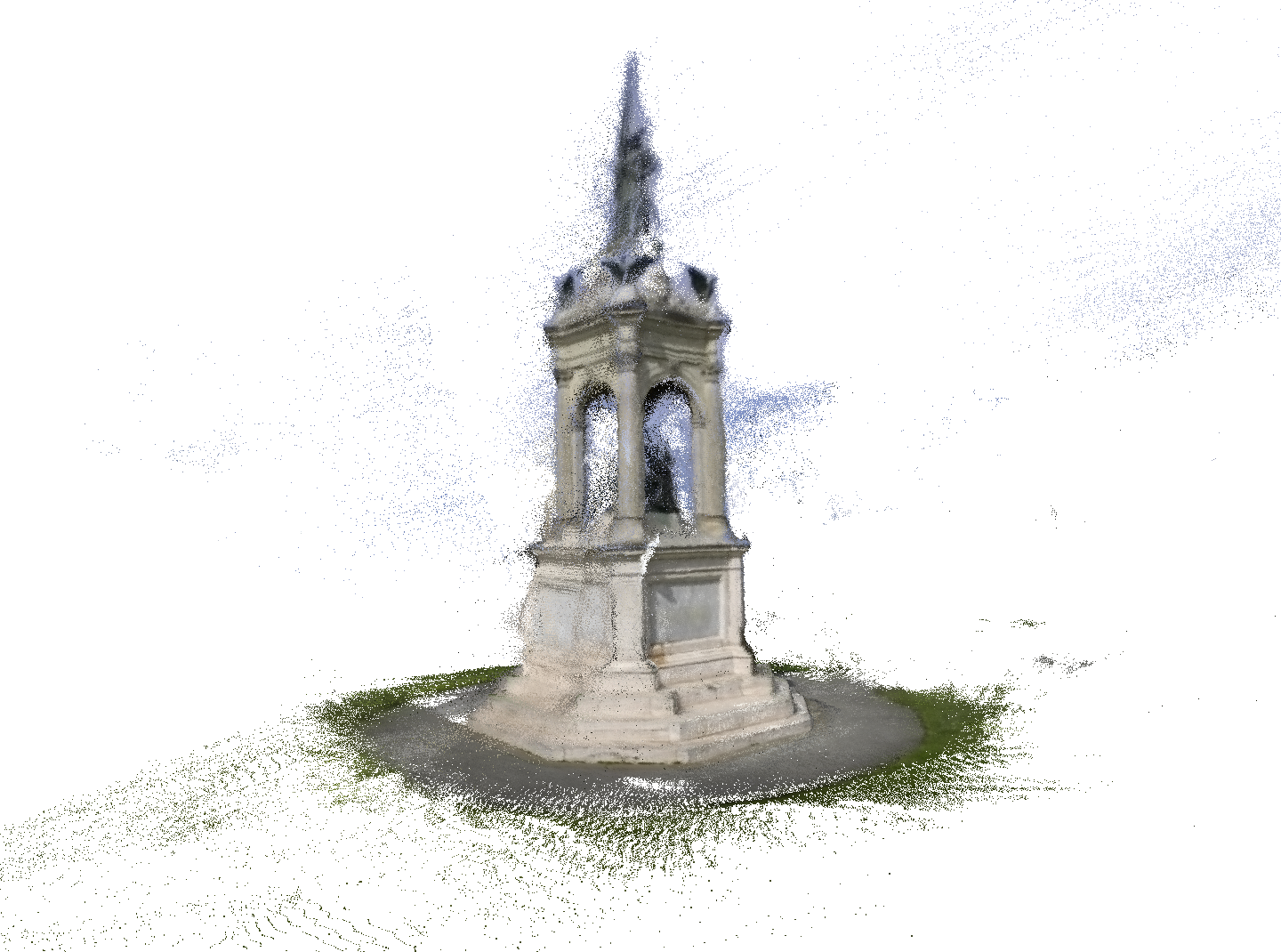}} &
\subfigure{
\includegraphics[width=0.2\linewidth]{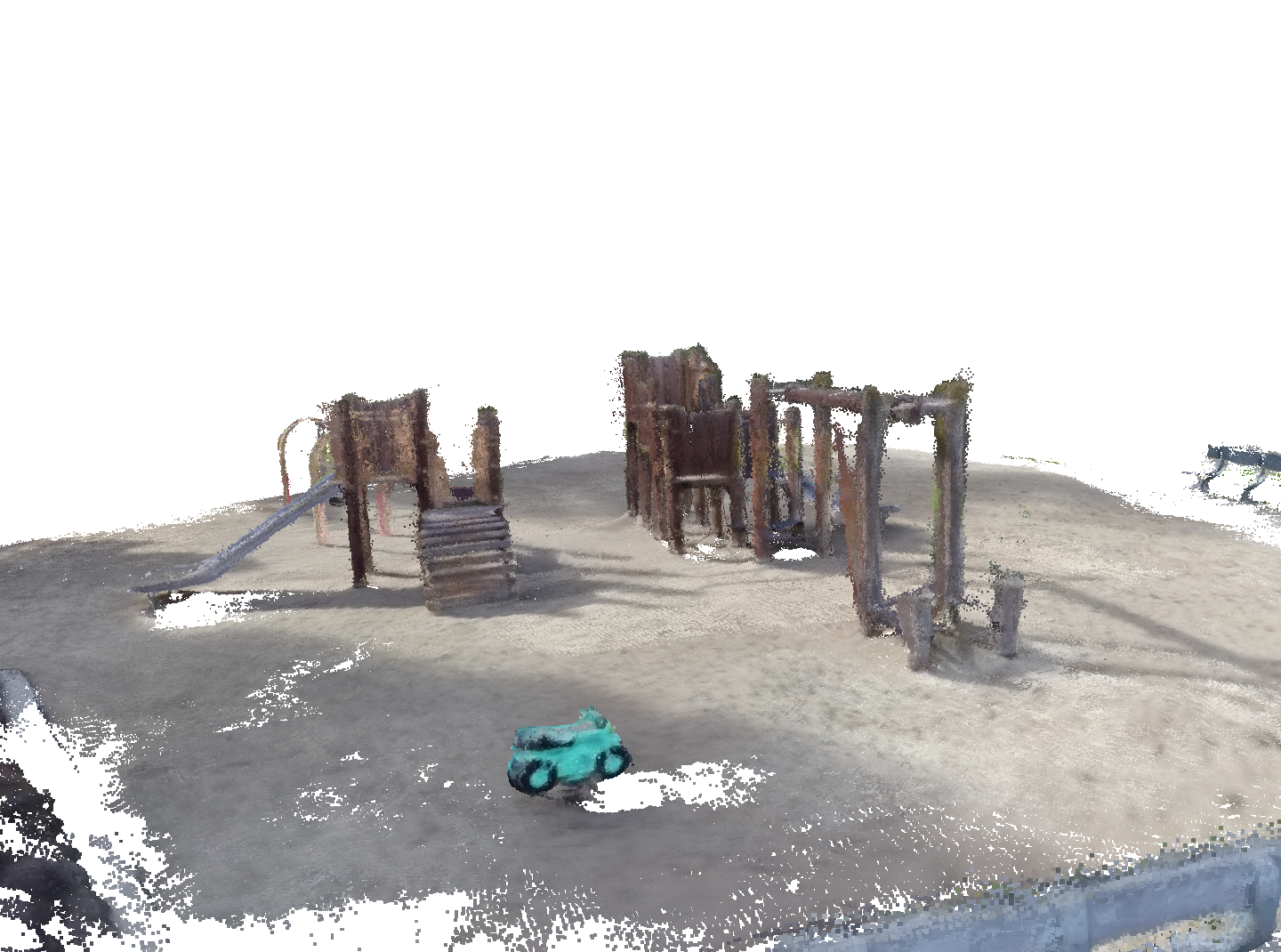}}\\
\footnotesize{(a) Train }  & \footnotesize{(b) Panther} & \footnotesize{(c) Francis}  & \footnotesize{(d) Playground} \\

\subfigure{
\includegraphics[width=0.2\linewidth]{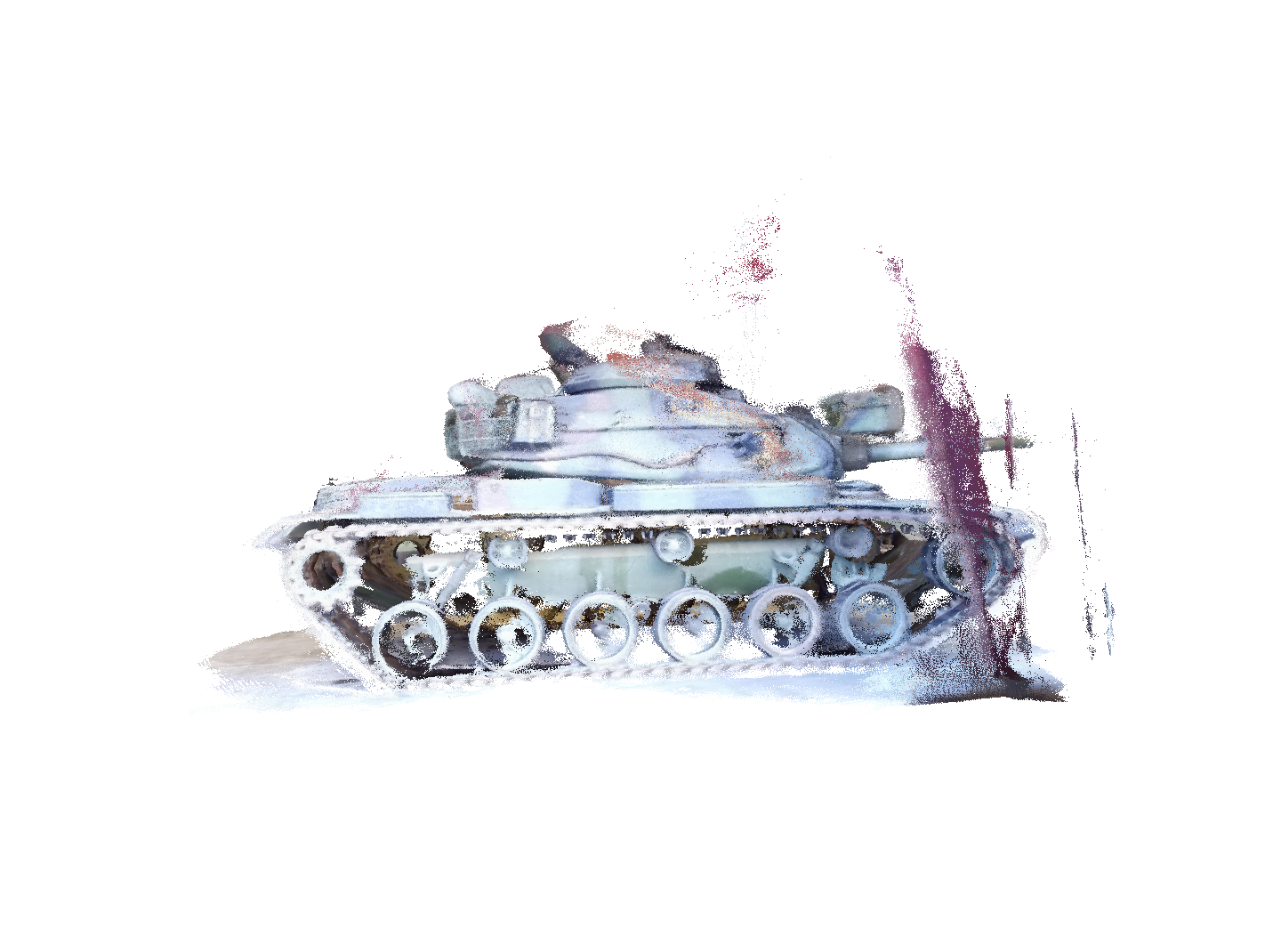}} &
\subfigure{
\includegraphics[width=0.2\linewidth]{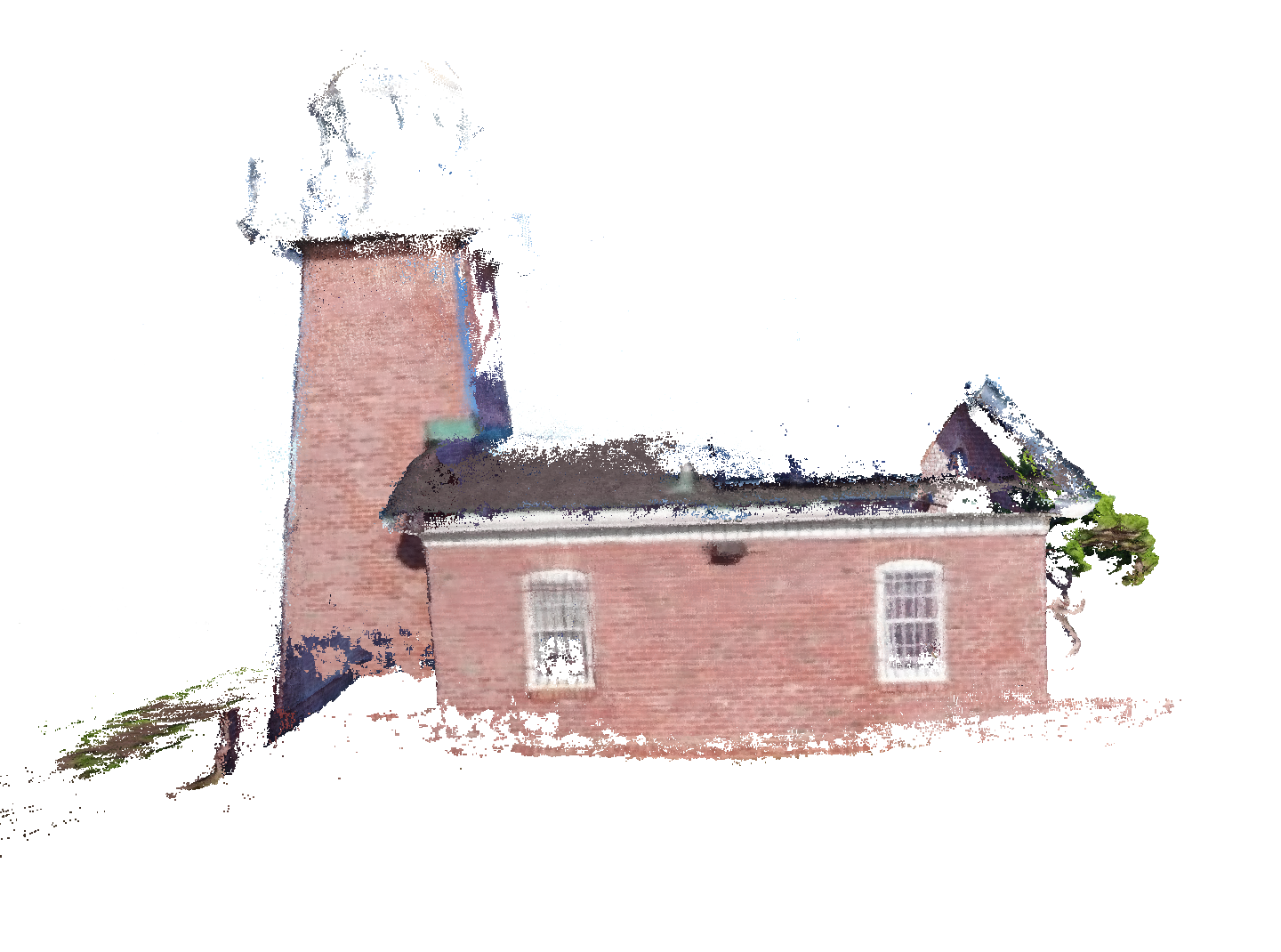}}&
\subfigure{
\includegraphics[width=0.2\linewidth]{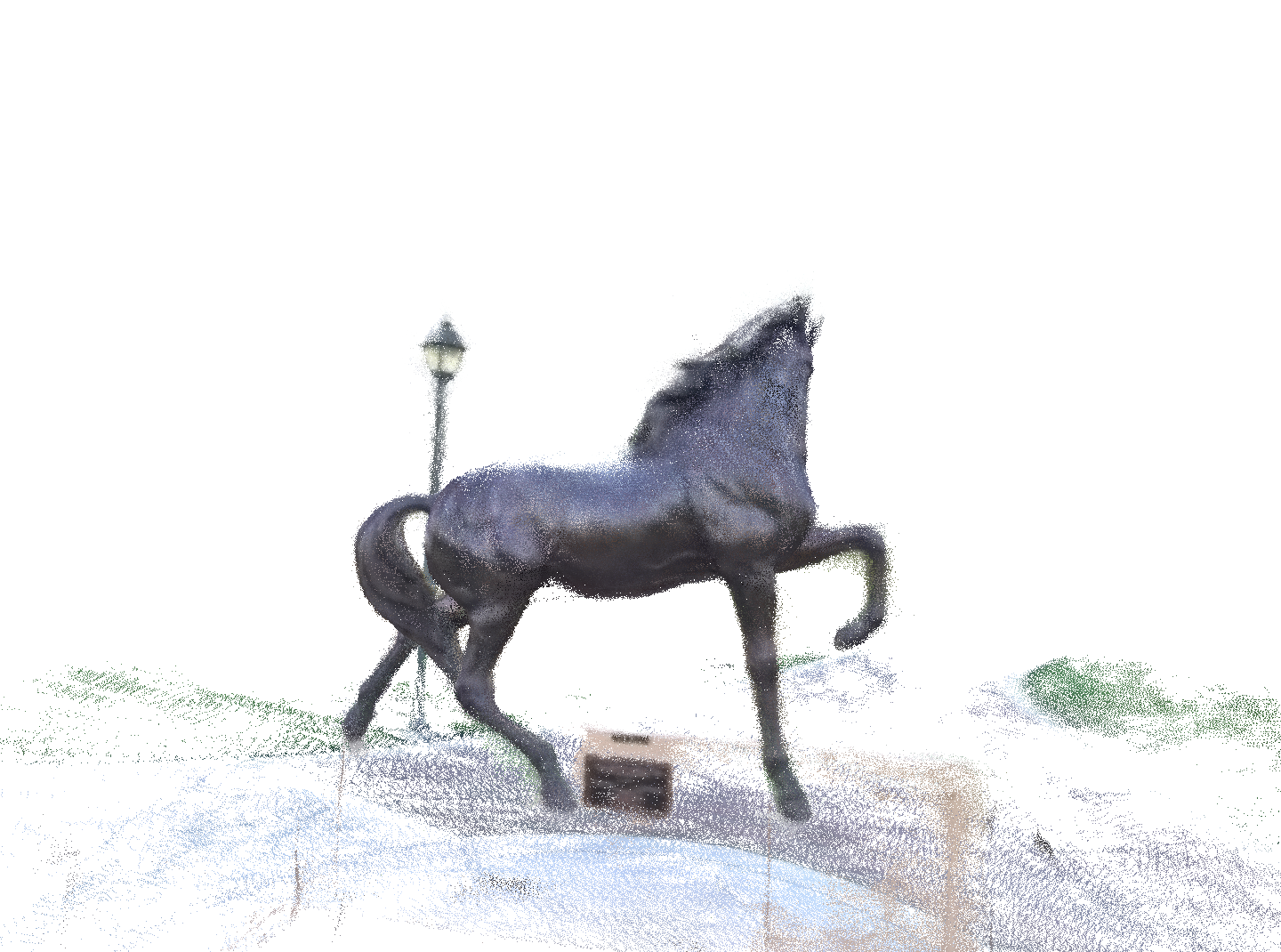}} &
\subfigure{
\includegraphics[width=0.2\linewidth]{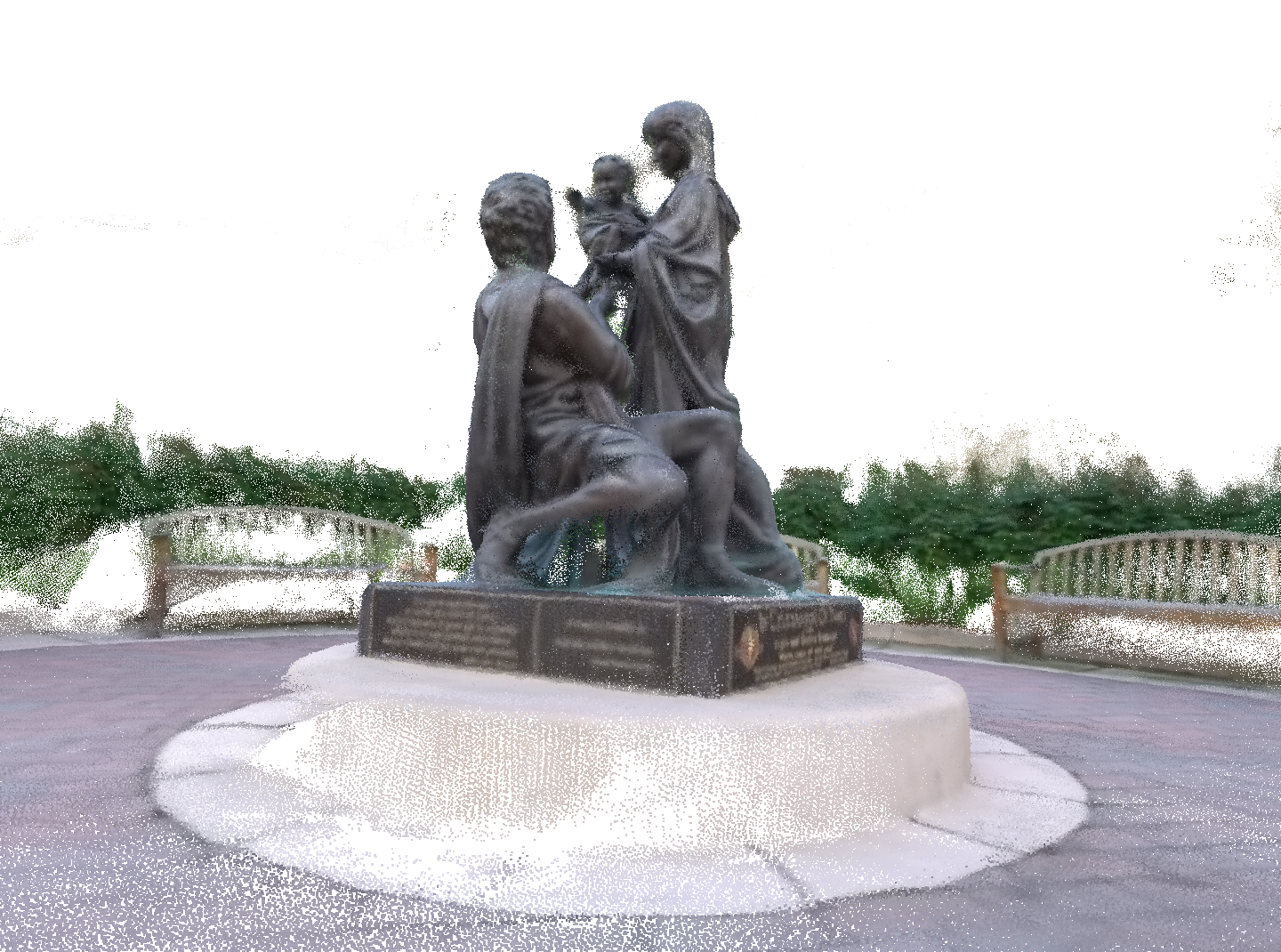}}\\
\footnotesize{(e) M60}  & \footnotesize{(f) Lighthouse} & \footnotesize{(g) Horse}  & \footnotesize{(h) Family} \\
\end{tabular}
\caption{\label{fig:Generalization points cloud results of MVS$^2$ on Tanks and Temples without any finetuning.}\textbf{3D point clouds generated by our MVS$^2$ without any finetuning on the Tanks and Temples dataset.}}
\end{center}
\end{figure*}

\begin{figure*}[!htp]
\begin{center} 
\subfigure{
\includegraphics[width=0.15\linewidth]{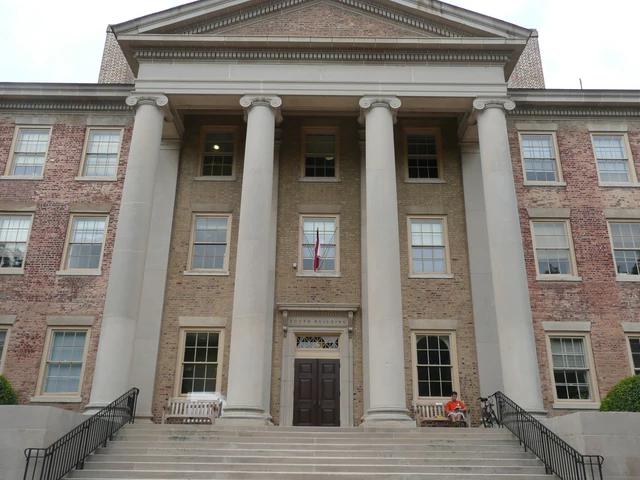} }
\subfigure{
\includegraphics[width=0.15\linewidth]{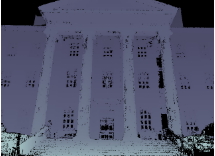} } 
\subfigure{
\includegraphics[width=0.15\linewidth]{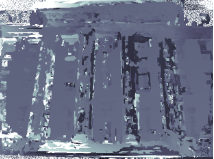} } 
\subfigure{
\includegraphics[width=0.15\linewidth]{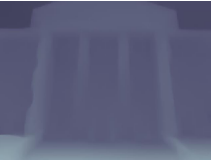} }
\subfigure{
\includegraphics[width=0.15\linewidth]{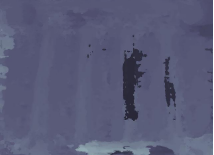} }
\subfigure{
\includegraphics[width=0.15\linewidth]{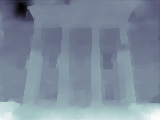} } 


\subfigure{
\includegraphics[width=0.15\linewidth]{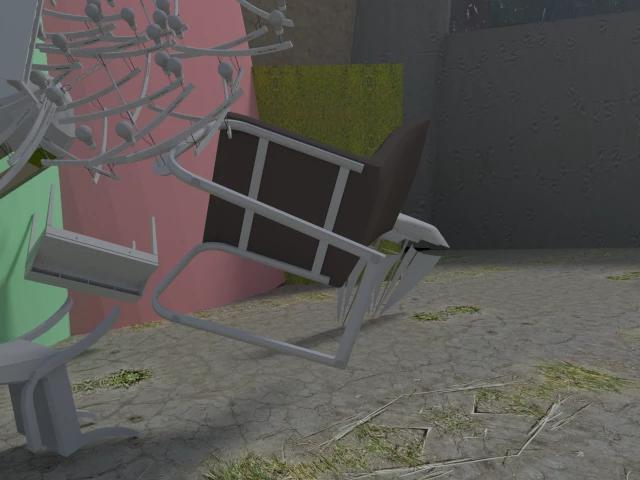} }
\subfigure{
\includegraphics[width=0.15\linewidth]{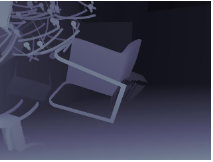} } 
\subfigure{
\includegraphics[width=0.15\linewidth]{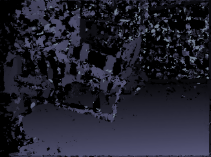}}
\subfigure{
\includegraphics[width=0.15\linewidth]{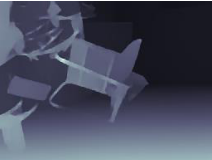} } 
\subfigure{
\includegraphics[width=0.15\linewidth]{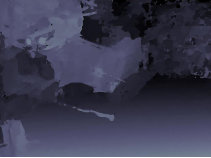} } 
\subfigure{
\includegraphics[width=0.15\linewidth]{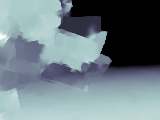} } 

\subfigure{
\includegraphics[width=0.15\linewidth]{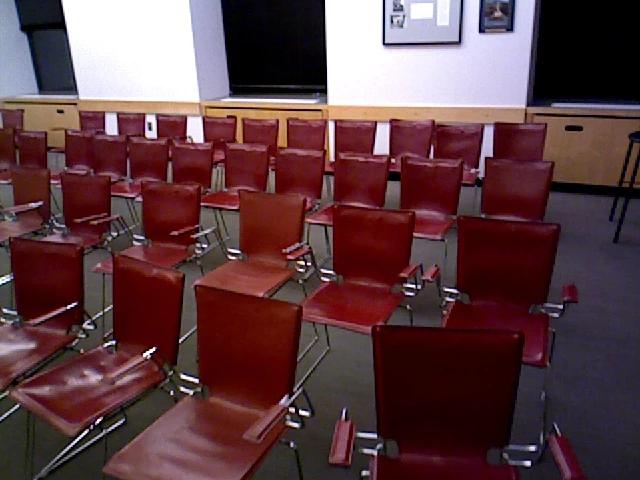} }
\subfigure{
\includegraphics[width=0.15\linewidth]{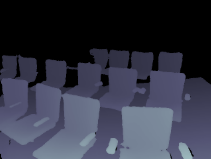} }
\subfigure{
\includegraphics[width=0.15\linewidth]{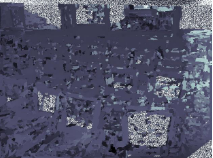} }
\subfigure{
\includegraphics[width=0.15\linewidth]{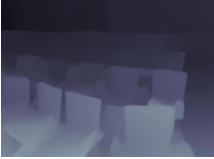} }
\subfigure{
\includegraphics[width=0.15\linewidth]{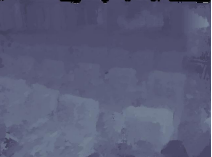} }
\subfigure{
\includegraphics[width=0.15\linewidth]{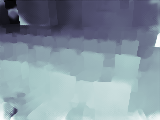} }
\caption{\label{fig:Comparsion visuslize results with other supervised methods}\textbf{Performance comparison on depth maps estimation with other supervised methods:} From Left to Right: reference image, ground truth depth map, depth map of COLMAP, depth map of DeMoN, depth map of DeepMVS and our result.}
\end{center}
\end{figure*}

\subsection{Generalization Ability}
As agreed in monocular depth estimation and binocular stereo matching, the supervised depth estimation methods strongly depend on the availability of large scale ground truth 3D data and the generalization ability could be hindered when evaluated on never-seen-before open-world scenarios. Here, we would like to verify the generalization ability of our unsupervised MVS network model. We conducted experiments on SUN3D, RGBD, MVS and Scenes11 datasets using our pre-trained model without any fine tuning. In Table~\ref{tab:The Excellent Generalization Ability}, we compare the performance of our MVS$^2$ with state-of-the-art traditional MVS methods and supervised MVS methods. We can conclude from Table \ref{tab:The Excellent Generalization Ability} that: 1) Our MVS$^2$ outperforms state-of-the-art traditional geometry-based multi-view method COLMAP\cite{Sch2016Pixelwise} with a wide margin, which shows the benefits in exploiting the large scale datasets; 2) Compared with supervised MVS methods trained on each dataset individually, our MVS$^2$, even only trained on the DUT training dataset, outperforms current state-of-the-art supervised MVS method DeepMVS on part of the error metrics. Qualitative comparison between our MVS$^2$ and competing MVS methods (COLMAP, DeMoN, DeepMVS) on the RGBD dataset is demonstrated in Fig.~\ref{fig:Comparsion visuslize results with other supervised methods}, where our method consistently achieves compared performance with SOTA supervised methods.

We also conducted experiments on the Tanks and Temples datasets without any fine tuning to validate the generalization ability of our network model. We choose $N=3$, $W=1920$, $H=1024$ and $D=192$ for our experiments. Qualitative point cloud results are presented in Fig.~\ref{fig:Generalization points cloud results of MVS$^2$ on Tanks and Temples without any finetuning.}, where our MVS$^2$ could reconstruct very detailed 3D structures.

\section{Conclusions}
In this paper, we have proposed the first unsupervised learning based MVS network, which learns the depth map for each view simultaneously without the need of ground truth 3D data. With our proposed multi-view symmetry network design, we can enforce the cross-view consistency of depth maps during training and testing. Our learned multi-view depth maps comply with the underlying 3D geometry. Our network learns multi-view occlusion maps in an alternative way. Experimental results on multiple benchmarking datasets demonstrate the effectiveness and excellent generalization ability of our network. In the future, we plan to extend the depth consistency beyond pairwise relation, such as consistency inside a clique. Extension to dynamic scenes \cite{CMU-MVS-Capture} could be another interesting future direction.

\section*{Acknowledgement}
{This research was supported in part by the Natural Science Foundation of China grants (61871325, 61420106007, 61671387). We thank all anonymous reviewers for their valuable comments.}

{\small
\bibliographystyle{ieee_fullname}
\bibliography{MVS_Reference}
}

\end{document}